%% file: neurips_2024.tex
\documentclass{article}

\usepackage[preprint]{neurips_2024}

\usepackage[utf8]{inputenc} 
\usepackage[T1]{fontenc}    
\usepackage{hyperref}       
\usepackage{url}            
\usepackage{booktabs}       
\usepackage{amsfonts}       
\usepackage{nicefrac}       
\usepackage{microtype}      
\usepackage[dvipsnames]{xcolor}         
\usepackage{graphicx}
\usepackage{subcaption}
\usepackage{multirow}
\usepackage{float}
\usepackage[most]{tcolorbox}
\usepackage{lscape}
\usepackage{adjustbox}
\usepackage{longtable}

\usepackage{algorithm}
\usepackage{algpseudocode}
\usepackage{amsmath}

\newcommand{\method}{{MeteoRA}}

\title{MeteoRA: Multiple-tasks Embedded LoRA\\ for Large Language Models}

\author{
Jingwei Xu$^{1}$\thanks{Corresponding author} \quad Junyu Lai$^1$ \quad Yunpeng Huang$^1$ \\
$^1$Department of Computer Science and Technology, Nanjing University\\
\texttt{jingweix@nju.edu.cn, \{junyu\_lai,hyp\}@smail.nju.edu.cn}
}

\begin{document}

\maketitle

\begin{abstract}
The \textit{pretrain+fine-tune} paradigm is foundational for deploying large language models (LLMs) across various downstream applications. Within this framework, Low-Rank Adaptation (LoRA) stands out for its parameter-efficient fine-tuning (PEFT), producing numerous reusable task-specific LoRA adapters. However, this approach requires explicit task intention selection, posing challenges for autonomous task sensing and switching during inference with multiple existing LoRA adapters embedded in a single LLM.
In this work, we introduce \textbf{\method} (\textbf{M}ultipl\textbf{e}-\textbf{t}asks \textbf{e}mbedded L\textbf{oRA}), a scalable and efficient framework that reuses multiple task-specific LoRA adapters into the base LLM via a full-mode Mixture-of-Experts (MoE) architecture.
This framework also includes novel MoE forward acceleration strategies to address the efficiency challenges of traditional MoE implementations.
Our evaluation, using the LlaMA2-13B and LlaMA3-8B base models equipped with 28 existing LoRA adapters through \method, demonstrates equivalent performance with the traditional PEFT method. Moreover, the LLM equipped with \method\ achieves superior performance in handling composite tasks, effectively solving ten sequential problems in a single inference pass, thereby demonstrating the framework's enhanced capability for timely adapter switching.
    
\end{abstract}

\section{Introduction}\label{sec:intro}
\input{sections/introduction}

\section{Background}\label{sec:back}
\input{sections/background}

\section{The Proposed MeteoRA}\label{sec:method}
\input{sections/method}

\section{Evaluation}\label{sec:experiment}
\input{sections/experiment}

\section{Related Work}\label{sec:relatedwork}
\input{sections/relatedwork}

\section{Limitations}\label{sec:limitation}
\input{sections/limitations}

\section{Conclusions}\label{sec:conclusion}
\input{sections/conclusion}


\newpage
\bibliographystyle{unsrtnat}
\bibliography{references}

\newpage
\appendix
\section{Appendix}\label{sec:appendix}
\input{sections/appendix}

\end{document}

%% file: sections/introduction.tex
Large language models (LLMs) have achieved significant advancement in modern intelligent applications, excelling in tasks from language comprehension to generation within the field of natural language processing (NLP)~\citep{achiam2023gpt, touvron2023llama}. By applying the fine-tuning process to pretrained LLMs, these models have demonstrated remarkable efficacy in handling domain-specific tasks. Examples include converting natural language text into SQL queries~\citep{katsogiannis2023survey, pourreza2024din}, utilizing LLMs as agents in diverse interactive applications~\citep{song2023llm, chen2023driving, gupta2023visual}, and developing models tailored for specific domains, such as BloombergGPT~\citep{wu2023bloomberggpt} for financial analysis and ChatLaw~\citep{cui2023chatlaw} for legal consulting.

This \textit{pretrain-fine-tune} paradigm has catalyzed the development of several parameter-efficient fine-tuning (PEFT) methods.
Low-Rank Adaptation (LoRA)~\citep{hu2021lora} stands out as a noteworthy exemplar of PEFT, offering efficient fine-tuning by updating only the low-rank matrices while keeping the rest of base LLM's parameters unchanged.
Once fine-tuned, these matrices, which consist of a minimal number of parameters, are encapsulated as a LoRA adapter that can be readily deployed or integrated with the base LLM for enhanced functionality.
To improve the capability of handling multiple tasks simultaneously,
the scalability of deploying these fine-tuned LoRA adapters has been explored. Solutions such as Huggingface PEFT~\citep{peft}, S-LoRA~\citep{sheng2023s}, and other variants have been developed to facilitate the simultaneous serving of numerous LoRA adapters on a single base LLM, enhancing the model's adaptability and efficiency in diverse application environments.
\begin{figure}[t]
	\begin{center}
	\includegraphics[width=0.8\columnwidth]{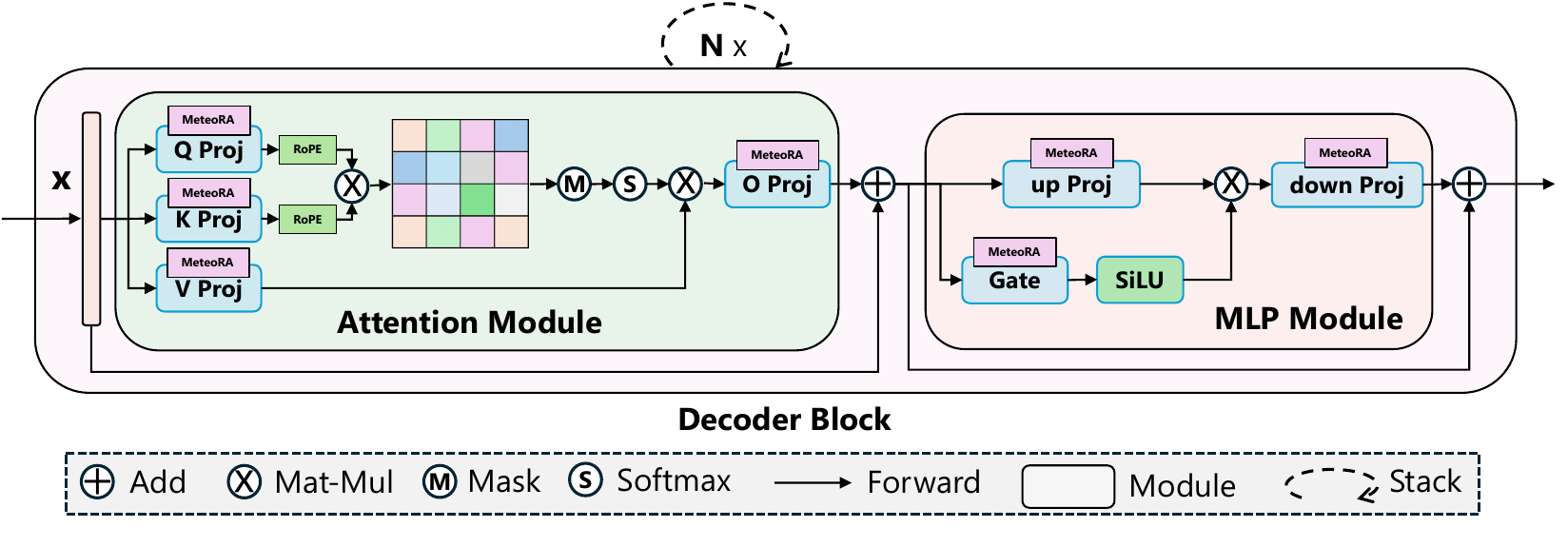}
		\caption{Our proposed framework provides a full-mode MoE architecture that directly reuses various off-the-shelf LoRA adapters, enhancing the LLM's ability to timely and autonomously activate appropriate adapters for the input. \method\ modules could be integrated into all basic linear layers of both Attention and MLP modules.  With the MoE forward acceleration strategies, LLM equipped with \method\ could be capable of addressing tasks across a wide range of domains effectively. \label{fig:framework}}
		\end{center}
\end{figure}

Despite the success of LoRA in the \textit{pretrain-fine-tune} paradigm, several challenges remain. When reusing existing LoRA adapters, a primary challenge is the ability of multi-LoRA embedded LLMs to autonomously and on-demand LoRA selection during inference, a process that should allow LLM to handle different tasks by activating the appropriate LoRA adapters without explicit user instructions. Furthermore, managing composite tasks that require timely switching between LoRA adapters presents difficulties, especially when these tasks involve multiple sub-problems each requiring specific adapter activation. Current approaches such as Huggingface PEFT and S-LoRA, while capable of serving multiple existing adapters simultaneously, mainly focus on loading rather than autonomously activating adapters, thus requiring manual intervention. Similarly, current LoRA fusion methods such as LoRAHub \citep{huang2023lorahub} and MoA \citep{feng2024mixture}, although they integrate and merge knowledge from various adapters, are not suitably designed to fuse a wide range of existing LoRA adapters with such a limited MoE framework, and lack the evidence in effectively managing dynamic adapter switching during inference for composite tasks.

In this paper, we introduce a novel multi-tasks embedded LoRA framework for LLMs to reuse existing LoRAs with the ability of autonomous task sensing and switching.
The framework proposes a MoE-style module called \method. Each \method\ module provides a trainable Gating network with MoE forward acceleration strategies (overcome the efficiency issue in naive MoE, especially when number of experts is much larger than 8) for all LoRAs' low-rank matrices in the linear layer. 
As shown in Figure \ref{fig:framework}, the \method\ module is applicable for all kinds of layers in Transformer-based LLMs (Q, K, V, and O in attention module and up\_proj, gating for SiLU \citep{elfwing2018sigmoid}, and down\_proj in MLP Module). Through fine-tuning all gates with minimal resources, \method\ effectively integrates the existing LoRA adapters into the base LLM model with the ability of autonomously on-demand LoRA selection, without the requirements of any explicit user or system instructions. Furthermore, the presence of numerous gates\footnote{For the LlaMA3-8B model, there are 224 \method\ modules in total, with each of the 32 decoder layers containing 7 gates (Q, K, V, O, up\_proj, gating, and down\_proj).} enhances the model with a full-mode MoE architecture, showing the capability of timely LoRA switching, addressing composite tasks with only two-shot examples as illustrations for all inputs.
Our empirical evaluations, which embedded 28 existing LoRA adapters with \method\ to LlaMA2-13B-base and LlaMA3-8b-base, highlight the full-mode MoE capabilities and demonstrate a significant performance enhancement. This improvement is particularly notable in handling composite tasks, showcasing the efficacy of the \method\ framework.
The primary contributions of \method\ are summarized as follows: 
\begin{itemize}
    \item \textbf{Scalable LoRA integration:} \method\ framework for reusing existing LoRA adapters advances the LLM's capability of autonomous on-demand LoRA selection and switching.

    \item \textbf{MoE forward acceleration:} revealing efficiency issue of MoE and providing the forward acceleration strategies with new GPU kernel operators to achieve a $\sim\!\mathbf{4 \times}$ speedup in average while maintaining memory overhead.

    \item \textbf{Advanced performance:} Evaluation shows superior performance in composite tasks when applying \method. thereby extending the practical utility of LLMs incorporating off-the-shelf LoRA adapters.
\end{itemize}

\noindent\emph{Organization.} Section~\ref{sec:back} discusses the background of LoRA adapters and the mixture-of-experts. Section~\ref{sec:method} formulates and details the proposed method for embedding numerous existing LoRA adapters into one base LLM. Section~\ref{sec:experiment} presents empirical evaluations. Section~\ref{sec:relatedwork} covers related work. Section~\ref{sec:limitation} discusses the limitations. Section~\ref{sec:conclusion} concludes the paper.

%% file: sections/background.tex
\textbf{Low-Rank adaption.}
Low-Rank Adaptation (LoRA)~\citep{hu2021lora} proposes a method to reduce the number of trainable parameters required for fine-tuning in downstream tasks. LoRA injects two trainable low-rank matrices $A \in \mathbb{R}^{d \times r}$ and $B \in \mathbb{R}^{r \times h}$ into each basic linear layer's weight matrix $W \in \mathbb{R}^{d \times h}$ of the Transformer-based LLM $\mathcal{M}$. The matrix multiplication of $A$ and $B$ represents the updates $\Delta W$ to the weight matrix $W$ when fine-tuning the model. The LoRA adapter modifies the forward process of this layer as follows:
\begin{equation}\label{eq:lora_forward}
\boldsymbol{o} = \boldsymbol{o}_{base} + \Delta\boldsymbol{o} = \boldsymbol{x}W_{base} + \boldsymbol{x}\Delta W = \boldsymbol{x}W + ((\boldsymbol{x}\times A) \times B)    
\end{equation}
where $\boldsymbol{x}\in \mathbb{R}^d$ represents the input hidden states for any token, $A, B$ first project it to the low-rank embedding space $\mathbb{R}^r$ and then map it back to the output space $\mathbb{R}^{h}$.
LoRA can be applied to seven types of linear layers in the Transformer: four in the self-attention module ($W_q$, $W_k$, $W_v$, and $W_o$) and three in the MLP module ($W_{\mathrm{up\_proj}}$, $W_\mathrm{gating}$, and $W_\mathrm{down\_proj}$).
Training LoRA adapters is straightforward. It continues to use the optimization target of causal language modeling to update LoRA's parameters while freezing the billions of parameters in the pretrained LLM $\mathcal{M}$.

\textbf{Multi-task LoRA fusion.}
LoRA adapter is usually fine-tuned to a specific downstream task.
To enhance the capacity of LLMs in handling multiple tasks, two paradigms are utilized in practice. 
One approach is to fuse datasets from different tasks and then fine-tune a single LoRA module on this combined dataset. However, \cite{ling2024domain} points out the difficulty in learning all specialized knowledge of various domains in one LLM.
The other approach leverages existing LoRA adapters as off-the-shelf components, directly merging these adapters into one base LLM.
Current popular LoRA frameworks, such as PEFT \citep{peft} and S-LoRA \citep{sheng2023s}, allow fusing multiple LoRA adapters. However, these frameworks must explicitly assign the active injected LoRAs, leaving an obvious disadvantage of lacking autonomous on-demand LoRA selection and timely LoRA switching during inference.
Existing work, such as LoRAHub \citep{huang2023lorahub}, could combine multiple LoRA adapters without the explicit task intention given by humans. However, few-shot/in-context learning is required for LoRAHub for every single downstream task. 

\textbf{Mixture-of-Experts.}
MoE is a machine learning paradigm that enhances model performance and efficiency by combining predictions from multiple specialized models, or experts. Introduced by \cite{jacobs1991adaptive}, MoE uses a gating network to assign input data to the most relevant experts dynamically. This approach leverages specialized knowledge from different experts, improving overall performance on diverse and complex tasks.
Recent progress, particularly by \cite{shazeer2017outrageously}, has demonstrated the effectiveness of MoE in large-scale neural networks. By using sparsely-gated MoEs, where only a subset of experts is activated for each input, computational efficiency is significantly increased without compromising model capacity. This has proven particularly useful in scaling Transformer-based architectures for various applications, such as Mixtral \citep{jiang2024mixtral}.

%% file: sections/method.tex
\subsection{MeteoRA architecture}
Given a base LLM $\mathcal{M}$ and $n$ existing LoRA adapters $\{L_1, L_2, \cdots, L_n\}$ that have already been fine-tuned with the distinct tasks $\{D_1, D_2, \cdots, D_n\}$ on $\mathcal{M}$ separately, our objective is to integrate the $n$ existing LoRA adapters into the base $\mathcal{M}$ via \method\ framework, resulting in a LoRA embedded model $\mathcal{M}_\mathrm{embed}$.
Figure~\ref{fig:framework} demonstrates the \method\ module complemented to each basic linear layer in LLM.
The reused LoRA adapters are off-the-shelf ones, available from open-source communities or have been fine-tuned for specific tasks, 
Each LoRA adapter $L_i$ contains a set of low-rank matrices $\{A_i,B_i\}$. \method\ furnishes each basic linear layer with a wide MoE architecture to embed the low-rank matrices provided by $n$ LoRA adapters.

\begin{figure}[t]
	\begin{center}
	\includegraphics[width=0.7\columnwidth]{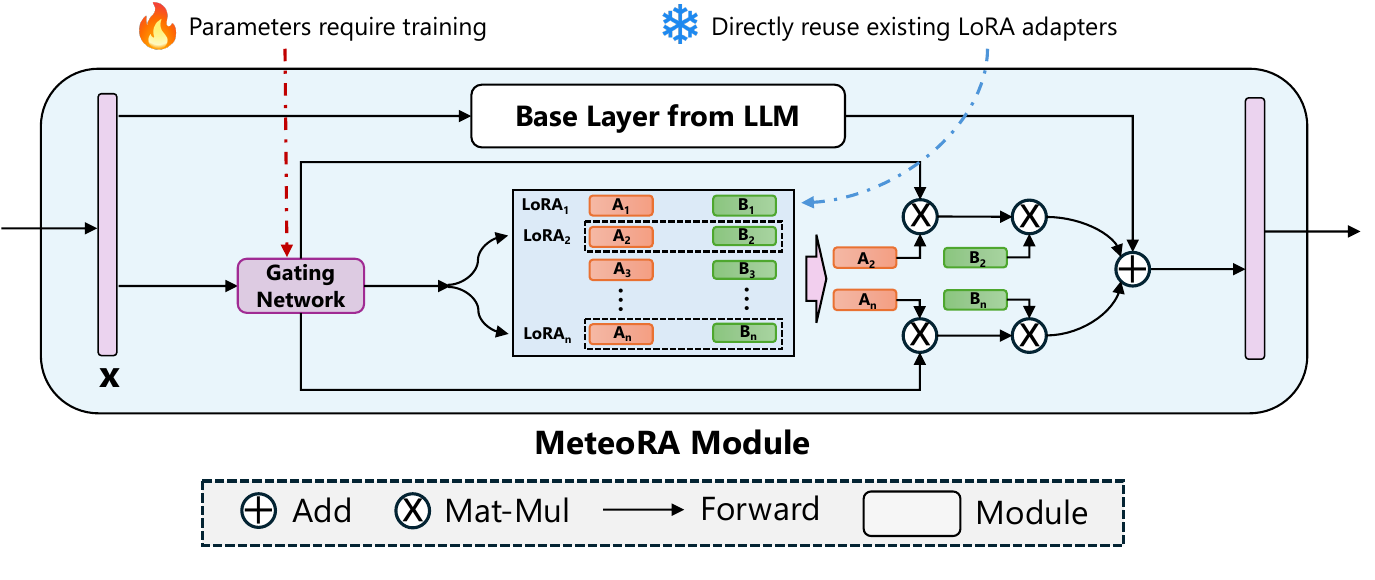}
		\caption{The architecture of \method\ module with MoE-style LoRA embedding. MeteoRA directly reuses existing LoRA adapters without fine-tuning and only requires training the Gating network.\label{fig:meteora}}
		\end{center}
\end{figure}

Figure~\ref{fig:meteora} shows the architecture of \method\ module. 
To embed $n$ existing LoRA adapters, \method\ module leverages the MoE architecture by injecting a trainable \textit{Gating network} $G: \mathbb{R}^{d}\rightarrow \mathbb{R}^{n}$ together with $n$ existing pairs of $\{A_i, B_i\}$ to $\mathcal{M}$. By applying $G(\boldsymbol{x})$, \method\ selects $k$ pairs of $\{A_i, B_i\}$ with the top-$k$ highest gated weights for each $x$. It then proceeds with the forward pass as follows:
\begin{equation}\label{eq:forward}
\boldsymbol{o} =\boldsymbol{o}_{base} + \Delta \boldsymbol{o}_{I(x)} = \boldsymbol{x}W_{\mathrm{base}} + \boldsymbol{x}\Delta W_{I(\boldsymbol{x})} = \boldsymbol{x}W_{\mathrm{base}} + \sum_{i\in I(\boldsymbol{x})}w_i \cdot( (\boldsymbol{x} \times A_i) \times B_i)
\end{equation}
where $I(\boldsymbol{x}):= \{i_1, i_2,..,i_k\}$ denotes the top-$k$ LoRAs selected for each token $\boldsymbol{x}$, which may varies from one another in every batch,
and $w_i$ is the normalized weight of the selected LoRA $L_i$.
$w_i$ could be calculated as follows:
\begin{equation}
w_i = \mathrm{softmax}(G_{i}(\boldsymbol{x}))=\frac{\exp(G_{i}(\boldsymbol{x}))}{\sum\limits_{j\in I(\boldsymbol{x})}\exp(G_{j}(\boldsymbol{x}))}
\end{equation}
where $G_i(\boldsymbol{x})$ denotes the unnormalized gated \textit{logits} for the $i$-th LoRAs. By doing this, the Gating network performs as a routing strategy for selecting the appropriate LoRA adapters based on the layer's input.
Each \method\ module contains a Gating network, and the Gating networks from different \method\ modules make decisions based on their own inputs, the selection of LoRA adapters could be dynamically switched in the forward process of each \method\ module through all LLM's decoder blocks.
\method\ also applies top-$1$ and top-$k$ gating strategies as detailed in Appendix \ref{appendix:topk}.

\subsection{Learning algorithm}
Training the injected \method\ modules adheres to the principles of fine-tuning LLM under autoregressive language modeling tasks. Given that $n$ pre-trained LoRA adapters, the training procedure for \method\ needs to maintain the parameters of the base LLM $\mathcal{M}$ and the $n$ LoRA adapters fixed.
Since \method\ supports top-$k$ experts (LoRAs) selection, we introduce the joint optimization that combines the loss of autoregressive language modeling $\mathcal{L}_{\mathrm{lm}}$ and all losses of Gating networks $\mathcal{L}_{\mathrm{gate}}$:
\begin{equation}\label{eq:loss}
\mathcal{L}=\mathcal{L}_{\mathrm{lm}}+\beta\mathcal{L}_{\mathrm{gate}}=\arg\max_{\theta}\sum_{i=1}^L(\log\mathrm{P}(x_i\mid x_{i-1};\theta)+\beta\sum_{j=1}^B\sum_{k=1}^ml_{k,j}(h))
\end{equation}
where $\beta$ is the hyper-parameter, $i$ is the token index, $L$ is the length of the language sequence represented as tokens, $x_i$ represents the token.
The loss $l_{k,j}$ is the cross-entropy loss for LoRA classification in one \method\ module.
For a base $\mathcal{M}$ contained $B$ decoder blocks with $m$ \method\ modules in each decoder, $\mathcal{L}_{\mathrm{gate}}$ sums the loss $l_{k,j}(h)$ based on the corresponding hidden inputs $h$.

\subsection{Forward acceleration}\label{sec:accelerate}
The core component of the \method\ module is a MoE architecture that incorporates $n$ existing LoRA adapters.
First of all, we allocate two big tensors $\mathcal{A}$ and $\mathcal{B}$ continuously on the HBM, each sized $(n, d, r)$ and $(n, r, h)$ to store all LoRA matrices $A_i$ and $B_i$, respectively. Then, as the real challenge, for each token $\boldsymbol{x}$ in the input sequence with the length $s$ for every batch with the size $b$, we should find its own candidate set $I(\boldsymbol{x})$ through $G$ to index $k$ pairs of $(A_i, B_i)$ from $\mathcal{A}$, $\mathcal{B}$ to apply the forward pass as Equation \ref{eq:forward}, making \method\ 
almost impossible to be as efficient as the \textit{single-lora} setting. Based on Mixtral~\cite{jiang2024mixtral}, the naive implementation, named \textit{loop-original}, might employ a for-loop to traverse $n$ LoRA adapters, and in the $i$-th iteration, it gathers the tokens $\{\boldsymbol{x} | i \in I(\boldsymbol{x})\}$ as a matrix $X_i \in \mathbb{R}^{p_i \times d}$, and then apply normal LoRA pass as $((X_i \times A_i) \times B_i)$. This method solves the selection problem by simply splitting the $b\times s$ tokens into $n$ sets, and do the forward pass sequentially for each set. However, considering the nature that $b\times s$ tokens are independent to each other, it can not fully take advantage of parallelized GEMM operators~\cite{gemm} by looping over all the $n$ adapters every time, especially when $p_i$ is really small (\textit{some adapters are only picked by few tokens}) or when $b\!\times\! s\! <\! n$ (like the \textit{auto-regressive} inference phase where $s$ is fixed to $1$), which might cost at most $\mathbf{10 \times}$ runtime compared to \textit{single-lora} for some tasks in our experiments (see Section \ref{sec:efcy}). 
 
The straightforward way to accelerate the forward pass, called \textit{bmm-torch}, is to directly index all the top-$k$ adapters for all the $b \times s$ tokens at the same time, leading to twice $bmm$ calculations followed by a batched weighted sum as:
\begin{equation}
    \underbrace{[\Delta\boldsymbol{o}_1, \Delta\boldsymbol{o}_2, .., \Delta\boldsymbol{o}_{bs}]}_{b\times s} =  \sum\limits_{k} \underbrace{[w_{1}, w..,w_{bsk}]}_{b\times s\times k} \odot \left(\big(\underbrace{[\boldsymbol{x}_1, .., \boldsymbol{x}_{bsk}]}_{b\times s\times k} \times \underbrace{[\boldsymbol{A}_{i_1}, .., \boldsymbol{A}_{i_{bsk}}]}_{b\times s\times k}\big) \times \underbrace{[\boldsymbol{B}_{i_1}, .., \boldsymbol{B}_{i_{bsk}}]}_{b\times s\times k}\right) \label{eq: bmm_torch}
\end{equation}

In contrast to \textit{loop-original}, \textit{bmm-torch} achieves $\sim\!\mathbf{4 \times}$ speedup by parallelizing all $b\!\times\! s \!\times\! k$ LoRA operations based on the PyTorch $bmm$ operator\cite{pytorch_bmm}, and only $\sim\!\mathbf{2.5 \times}$ slower than the upper bound \textit{single-lora} in most of our experiments (see Section~\ref{sec:efcy}). 

However, due to PyTorch's indexing constraints~\cite{pytorch_index}, \textit{bmm-torch} needs to allocate a new larger HBM by $\frac{b\!\times\! s \!\times\! k}{n}$ times to hold the batched $\mathcal{A}$, $\mathcal{B}$. Additionally, when $b$ or $s$ is quite large (\textit{like long-history multi-turn chat}), the large memory overhead of \textit{bmm-torch} will become a practical bottleneck. To address this, we develop a custom GPU kernel operator for the \method\ forward pass using Triton\cite{triton_lang}, which not only keeps the $80\!\%$ time efficiency with \textit{bmm-torch}, but also remains the low memory overhead at the \textit{loop-original} level (see Section~\ref{sec:efcy}). Details on this kernel operator are provided in Appendix~\ref{appendix: triton_kernel}.

%% file: sections/experiment.tex
We conduct experiments on individual and composite tasks as detailed in Section \ref{sec:eval_settings}.
For our base models, we use two well-known LLMs, LlaMA2-13B \citep{touvron2023llama} and LlaMA3-8B \citep{llama3}.
The code and the models are available\footnote{The implementation code is accessible at \url{https://github.com/ParagonLight/meteor-of-lora}, and the two \method\ embedded LLMs are available at\url{https://huggingface.co/ParagonLight/MeteoRA-llama2-13b} and \url{https://huggingface.co/ParagonLight/MeteoRA-llama3-8b}}.

\subsection{Evaluation settings}\label{sec:eval_settings}
\textbf{LoRA tasks and datasets.} We select 28 tasks from well-known benchmarks for our experiment. Specifically, our task set consists of 22 tasks from BigBench \citep{bigbench}, three non-English to English translation tasks from News-Commentary \citep{news_commentary}, and three widely utilized tasks: GSM8K \citep{gsm8k}, CNN/DailyMail \citep{cnn_dailymail}, and Alpaca \citep{alpaca}. These 28 tasks span a variety of NLP categories, such as contextual comprehension, conversational question answering, summarization, translation, mathematics, logical reasoning, and multilingual challenges. For detailed task descriptions, refer to Appendix \ref{appendix:28_task_details}.

\textbf{Metrics.} We apply a zero-shot evaluation setting for all tasks, adding brief task descriptions for tasks such as CNN/DailyMail and the three translation tasks that do not inherently include task descriptions.
As for metrics, we use \textit{accuracy} for multiple-choice tasks and GSM8K while employing metrics such as \textit{BLEU}, \textit{ROUGE-1}, \textit{ROUGE-2}, and \textit{ROUGE-L} for other tasks.

\begin{figure}[t]
    \centering
    \begin{subfigure}[b]{0.49\textwidth}
        \includegraphics[width=0.8\textwidth]
        {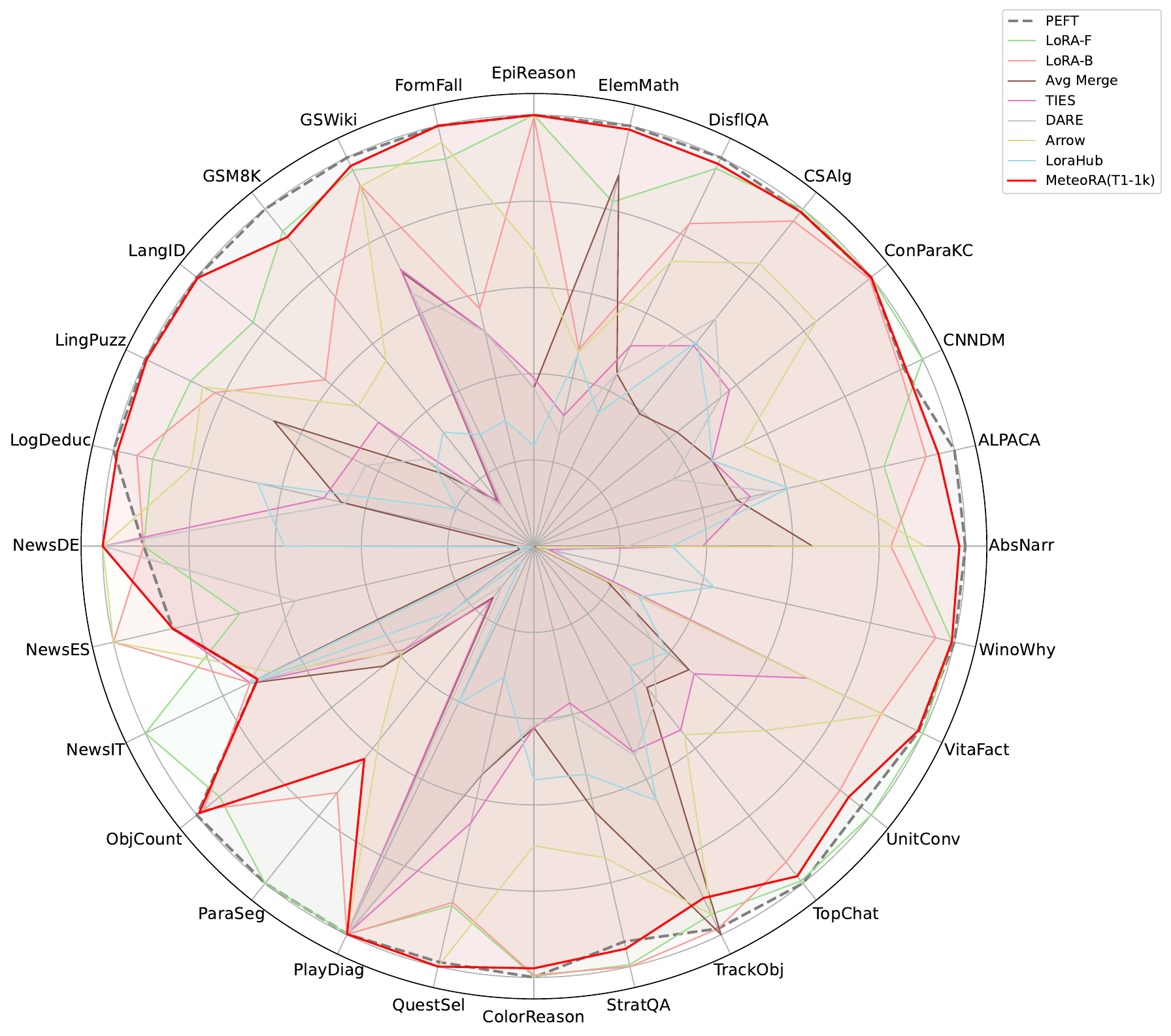}
        \caption{Evaluation results of models based on LlaMA2-13B.}
        \label{fig:llama2_13b_radar_graph}
    \end{subfigure}
    \hfill
    \begin{subfigure}[b]{0.49\textwidth}
        \includegraphics[width=0.8\textwidth]
        {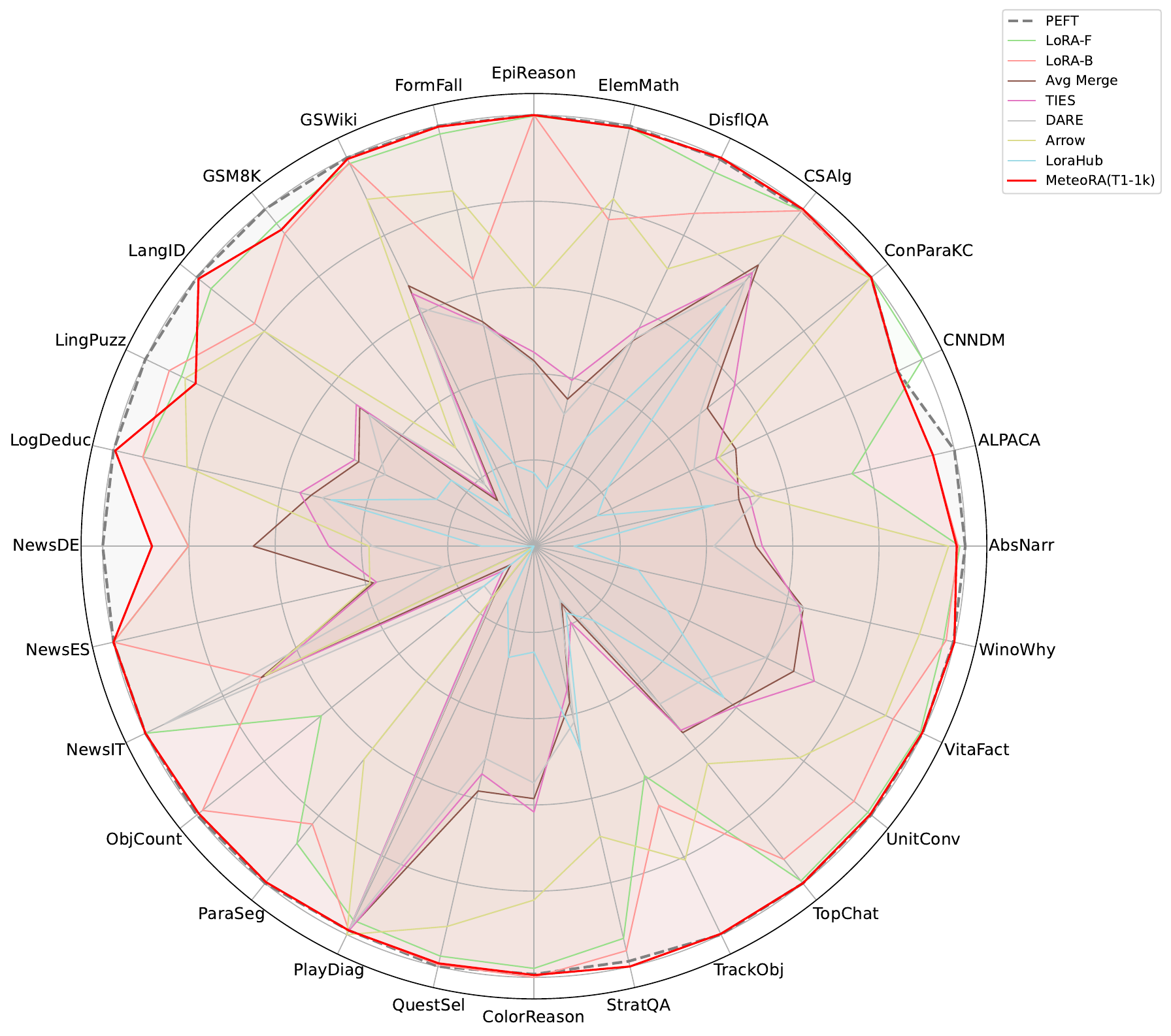}
        \caption{Evaluation results of models based on LlaMA3-8B.}
        \label{fig:llama3_8b_radar_graph}
    \end{subfigure}
    \caption{Evaluation results on the 28 selected tasks. The MeteoRA performs similarly on most tasks, leading to high overlap between the two polygons in the radar graphs. For clarity, we only draw results from MeteoRA with top-1 strategy in the radar graphs. Detailed results for each individual task are available in Appendix \ref{apppendix:28_task_results}}
    \label{fig:28_tasks_radar_graph}
\end{figure}
\textbf{Models.} We use LlaMA2-13B and LlaMA3-8B as the base LLMs for LoRA and \method\ adaption. Both LlaMA models are pretrained LLMs and do not include the process of instruction tuning. We train specific LoRA adapters for each task using their respective training sets. The process of training LoRA adapters could be offline or dismissed when off-the-shelf LoRA is accessible. Then, the Gating networks, which embed the adapters in the \method\ module, are fine-tuned efficiently based on the balanced dataset containing 1,000 samples for each task. The Gating networks for 28 tasks take no more than 10 hours to reach the convergence with 4 H800 training via Accelerate \citep{accelerate}. For scenarios where the training data for the original LoRA adapter is limited, we train Gating networks using a top-2 strategy, with only 100 and 5 samples accessible per task.

For baseline comparisons, we train one LoRA adapter (i.e., LoRA-F) using a mixed training set from all 28 tasks, and another LoRA adapter (i.e., LoRA-B) with the balanced dataset designed for training the Gating network. 
We also use Huggingface PEFT (short in PEFT) loading all 28 LoRA adapters (same ones used for \method) with explicit LoRA activation information during evaluation as a reference model. Additionally, we include several LoRA merge methods for comparison, including: averaging 28 LoRA adapters (referred to as Avg Merge), TIES \citep{ties}, DARE \citep{dare}, Arrow \citep{arrow}, and LoraHub \citep{huang2023lorahub}.

All LoRA adapters interact with all seven linear layers in LLaMA’s Decoder layer, configured with $r=8$, $\alpha=16$, and a learning rate of $5e-5$. Due to some tasks having small training sets, the batch size for fine-tuning is set to $4$. All our experiments were conducted on a GPU server with five H800 80G GPUs. Notice that we carefully selected the training hyperparameters for the LoRA-F and LoRA-B to ensure that their performance on the 28 tasks would not be excessively incomparable.

\textbf{Composite tasks.} To evaluate the model's capability of sequentially solving composite tasks, we construct three composite evaluation sets by serially concatenating independent tasks. These evaluation sets, referred to as \textit{composite-3}, \textit{composite-5}, and \textit{composite-10}, consist of 3, 5, and 10 tasks, respectively, each containing 200 samples. Models are expected to sequentially generate both the question number and the answer in the order of the tasks. Temperature scaling is involved in Gating network. More details refer to Figure \ref{fig:serial_3_short}, Appendix \ref{sec:serial_result_details} and \ref{appendix:serial_3}.

\subsection{Main results}\label{sec:ondemand}

Figures \ref{fig:llama2_13b_radar_graph} and \ref{fig:llama3_8b_radar_graph} demonstrate the performance of the \method\ models, LoRA-F, LoRA-B, 5 LoRA merge methods, and a reference model PEFT based on LlaMA2-13B and LlaMA3-8B, respectively, across the selected 28 tasks. Table \ref{tab:28_tasks_llama_avg} shows the averaged scores in various matrics for all methods.

\begin{table}[t]
\centering
\caption{Results of the 28 selected tasks on LlaMA2-13B/LlaMA3-8B base LLMs. T1 and T2 represent the top-1 and top-2 strategies, while the subsequent numbers indicate the number of accessible samples per task for gate training.
Our methods perform the best in most tasks. Notice that the task \textit{linguistics\_puzzles} achieves significantly higher ROUGE scores on LlaMA3-8B base, disproportionately influencing the average ROUGE scores and resulting in slightly higher averages for LoRA-B. Excluding this outlier, our methods consistently lead in performance across the evaluation.}
\label{tab:28_tasks_llama_avg}
\begin{tabular}{cccccc}
\toprule
Model           & Accuracy$\uparrow$ & BLEU$\uparrow$  & ROUGE-1$\uparrow$ & ROUGE-2$\uparrow$ & ROUGE-L$\uparrow$ \\
\color{gray}{PEFT (reference)}            & \color{gray}{0.762 / 0.817}    & \color{gray}{35.66 / 45.32} & \color{gray}{0.340 / 0.341}   & \color{gray}{0.163 / 0.164}   & \color{gray}{0.316 / 0.317}   \\
\midrule
LoRA-F         & 0.730 / 0.767    & 41.27 / 42.93 & 0.318 / 0.327   & 0.136 / 0.157   & 0.294 / 0.306   \\
LoRA-B         & 0.666 / 0.750    & 37.98 / 38.47 & 0.314 / \textbf{0.343}   & 0.128 / \textbf{0.171}   & 0.288 / \textbf{0.321}   \\
Avg Merge   & 0.370 / 0.427  & 19.23 / 39.89  & 0.231 / 0.200  & 0.082 / 0.060  & 0.184 / 0.158  \\
TIES       & 0.388 / 0.441  & \textbf{47.28} / 34.66  & 0.195 / 0.199  & 0.055 / 0.059  & 0.151 / 0.158  \\
DARE       & 0.332 / 0.404  & 46.531 / 36.74  & 0.192 / 0.188  & 0.054 / 0.056  & 0.144 / 0.147  \\
Arrow      & 0.569 / 0.647  & 41.03 / 29.93  & 0.281 / 0.283  & 0.123 / 0.142  & 0.234 / 0.242  \\
LoraHub    & 0.307 / 0.235  & 13.43 / 10.11  & 0.158 / 0.141  & 0.049 / 0.035  & 0.124 / 0.104  \\
\midrule
MeteoRA (T1-1k) & 0.755 / \textbf{0.811}    & 36.73 / \textbf{45.64} & \textbf{0.336} / 0.338   & 0.160 / 0.158   & 0.313 / 0.314   \\
MeteoRA (T2-1k) & 0.748 / 0.806    & 38.97 / 44.98 & \textbf{0.336} / 0.337   & \textbf{0.161} / 0.158   & \textbf{0.314} / 0.313   \\
MeteoRA (T2-100)     & \textbf{0.758} / 0.783  & 39.44 / 39.90  & 0.331 / 0.309  & 0.159 / 0.139  & 0.281 / 0.256  \\
MeteoRA (T2-5)       & 0.740 / 0.773  & 38.37 / 40.12  & 0.328 / 0.299  & 0.156 / 0.131  & 0.277 / 0.246  \\
\bottomrule
\end{tabular}
\end{table}

The evaluation results indicate that, regardless of the base LLM, the \method\ models utilizing the top-1 strategy achieve performance very close to the reference model PEFT, while no explicit LoRA activation/deactivation is required in \method. Although LLMs with both LoRA-F and LoRA-B reach comparable performance on several certain tasks, they exhibit significantly poorer outcomes on others. Additionally, \method\ employing the top-2 strategy, despite occasionally showing greater capability loss compared to \method\ with top-1 strategy, occasionally outperforms PEFT with adapters trained directly on the individual tasks. This suggests that the $L_{\mathrm{lm}}$ component in the loss function (Equation \ref{eq:loss}) becomes influential in these cases, indicating a beneficial mix of LoRA adapters from various tasks for future study. For the MeteoRA (T2-100) and MeteoRA (T2-5), although their performance shows a gap compared to MeteoRA 1k, they still outperform the baseline models on most metrics. This demonstrates that the Gating network can still learn to effectively utilize existing LoRA adapters with only a few examples.

\subsection{Composte-n tasks}\label{sec:composite}
The evaluation results for these three tasks are illustrated in Table \ref{tab:serial_eval_results}. 
Notice that only LlaMA3-8B with the \method\ (top-2 strategy) and LoRA-B effectively address these \textit{composite-n} tasks. Subsequent discussions will therefore focus exclusively on these two models.
Although the \method\ model attempts slightly fewer questions than LoRA-B in \textit{composite-3} tasks, it correctly answers a higher number of multiple-choice questions and achieves superior BLEU and ROUGE scores. As the task complexity increases to \textit{composite-5} and \textit{composite-10}, \method\ outperforms LoRA-B in almost all metrics. For more details, refer to Appendix \ref{sec:serial_result_details}.

\begin{figure}[t]
	\begin{center}
	\includegraphics[width=0.8\columnwidth]{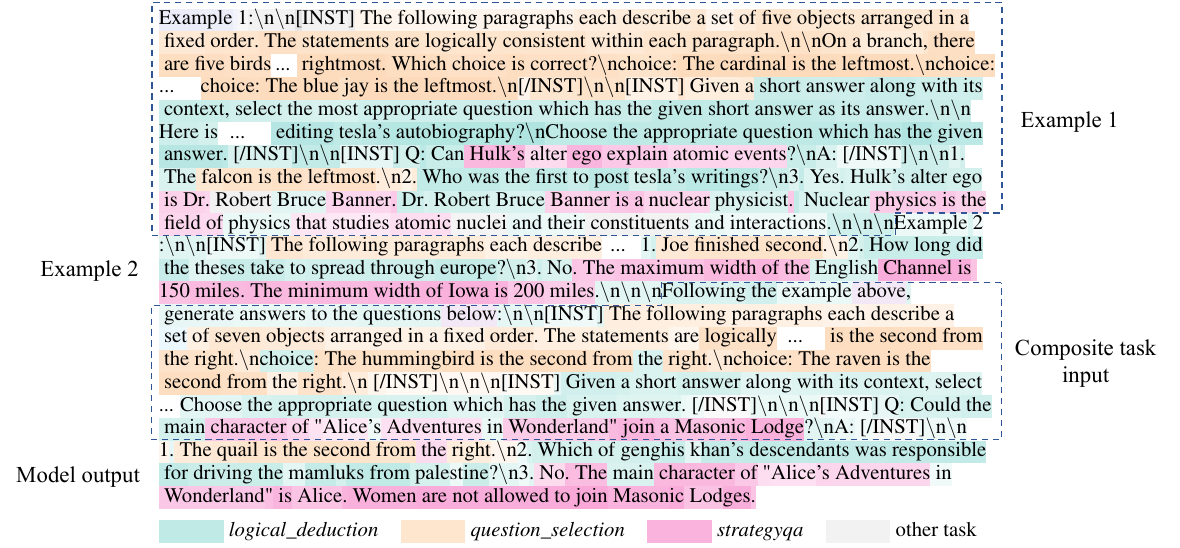}
		\caption{An example of \textit{composite-3} task. We highlight the statistically dominant LoRA selected by \method\ in token level (decoded to words). The result shows that LLM with \method\ could achieve timely LoRA switching on both phases of input understanding and output generation. The background color gets darker when Gating network assigns a higher weight value. \label{fig:serial_3_short}}
		\end{center}
\end{figure}

To further validate the functionality of the Gating network in the \method\ block, we display the LoRA selection patterns in the inference process of a \textit{composite-3} sample in Figure \ref{fig:serial_3_short}. With the top-$2$ strategy, Gating network appropriately assigns greater weight to the corresponding LoRA adapters for the majority of the tokens, no matter in input or output. At the junctions of two adjacent tasks, the Gating network correctly performed the timely switching actions of LoRA adapters.

\begin{table}[t]
\centering
\caption{The evaluation results of \textit{composite-n} tasks. MeteoRA is marked in color on the left side, while LoRA-B is in black on the right side. Refer to Appendix \ref{sec:serial_result_details} for a detailed explanation.}
\label{tab:serial_eval_results}
\begin{tabular}{ccccccc}
\toprule
Metric & \multicolumn{2}{c}{composite-3} & \multicolumn{2}{c}{composite-5} & \multicolumn{2}{c}{composite-10} \\
\midrule
\# Avg Attempt  & \textcolor{Red}{2.95}$\downarrow$  & 3.00  & \textcolor{Green}{4.63}$\uparrow$  & 4.33  & \textcolor{Green}{8.24}$\uparrow$  & 6.07  \\
\# Avg Correct & \textcolor{Green}{1.49}$\uparrow$  & 1.31  & \textcolor{Green}{2.62}$\uparrow$  & 2.42  & \textcolor{Green}{3.75}$\uparrow$  & 2.95  \\
Avg BLEU        & \textcolor{Green}{15.31}$\uparrow$ & 10.55 & \textcolor{Green}{9.86}$\uparrow$  & 9.41  & \textcolor{Green}{8.85}$\uparrow$  & 8.71  \\
Avg ROUGE-1     & \textcolor{Green}{0.195}$\uparrow$ & 0.135 & \textcolor{Green}{0.221}$\uparrow$ & 0.219 & \textcolor{Green}{0.238}$\uparrow$ & 0.161 \\
Avg ROUGE-2     & \textcolor{Green}{0.052}$\uparrow$ & 0.027 & \textcolor{Green}{0.069}$\uparrow$ & 0.063 & \textcolor{Green}{0.059}$\uparrow$ & 0.043 \\
Avg ROUGE-L     & \textcolor{Green}{0.182}$\uparrow$ & 0.128 & \textcolor{Red}{0.207}$\downarrow$ & 0.208 & \textcolor{Green}{0.209}$\uparrow$ & 0.123 \\
\bottomrule
\end{tabular}
\end{table}

\subsection{Efficiency}\label{sec:efcy}
\begin{figure}[t]
	\begin{center}
	\includegraphics[width=0.7\columnwidth]
    {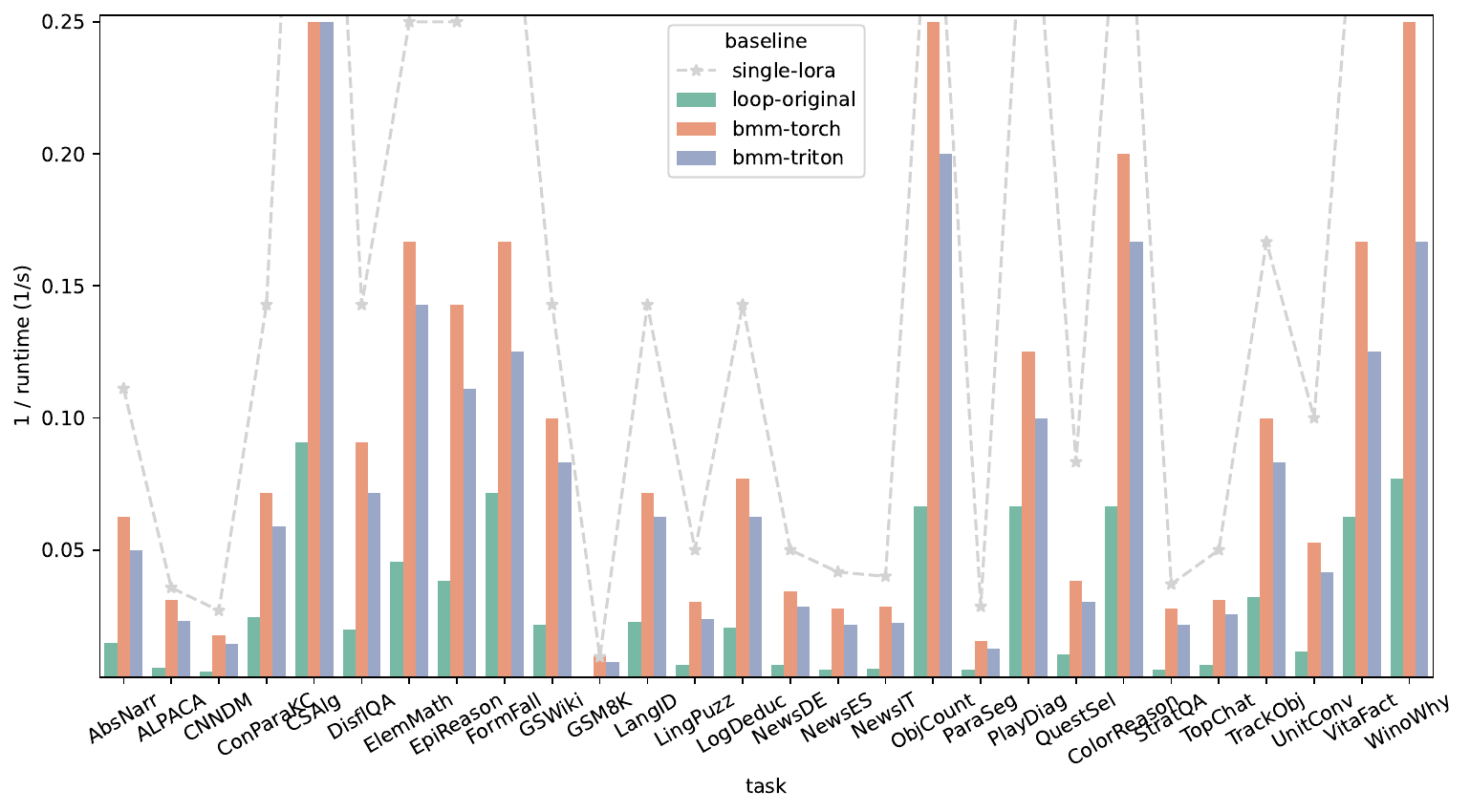}
		\caption{The overall \textit{root-of-runtime} of \textit{four} forward pass designs on 28 different Big-Bench subtasks. \label{fig:task-root-time}}
		\end{center}
\end{figure}

To assess the efficiency of our novel forward pass designs using custom GPU kernel operators, we truncate $batch\_size \!\times\! 10$ samples from each test dataset of all 28 tasks. We evaluate these designs alongside \textit{four} variants with the same hyperparameters: the upper-bound \textit{single-lora}, the baseline \textit{loop-original}, and two novel forward acceleration strategies based on $bmm$: \textit{bmm-torch} and \textit{bmm-triton}, implemented by PyTorch and Triton respectively. Figure~\ref{fig:task-root-time} displays the histogram of the overall \textit{root-of-runtime} metric for each task and design. Additional evaluation is detailed in Appendix~\ref{appendix: efcy_exps}.

%% file: sections/relatedwork.tex
{\bf Multi-task fusion.} Our proposed method falls into the field of LoRA adapter composition for multi-task fusion.
The first category focuses on fusing the entire models.
Researchers mainly study model ensembling and multi-task learning to achieve this goal. Existing works integrate the models under the setting of shared model architecture \citep{matena2022merging,jin2022dataless,wu2023pi,yadav2024ties}. Others focus on merging models with various architectures or from different tasks. Both methods \citep{stoica2023zipit, liu2022deep} try to merge models that are trained for various tasks without additional training.
The second category is more concerned with fusion in terms of the tasks.
\cite{ilharco2022editing} proposes a model editing method via task vectors.
\cite{sun2022multitask} leverage in-context learning with few-shots to enhance the performance of unseen tasks.
However, these methods require multi-task training or prior knowledge for the evaluation tasks.
Our method embeds off-the-shelf LoRA adapters with a Gate network in the \method\ module. None of the examples (zero-shot) are required for all individual tasks.

{\bf Fusion under MoE}. In the context of \textit{pretrain-fine-tune} paradigm, PEFT becomes a common sense for developing Transformer-based LLM downstream applications.
Directly fine-tuning on a fused dataset from various tasks is unable to achieve better performance \cite{ling2024domain}.
Some works focus on leveraging existing LoRA adapters as off-the-shelf components, integrating them directly into a base LLM. For example, PEFT \citep{peft} and S-LoRA \citep{sheng2023s} are frameworks aiming to embed multiple LoRA adapters to one LLM. However, requiring explicit activation/deactivition during usage.
MixLoRA \citep{li2024mixlora} targets to a resource-efficient sparse MoE model, fine-tuning MoE on MLP module with the auxiliary load balance loss used in Mixtral \citep{jiang2024mixtral}. Although MixLoRA supports LoRA adapters for the attention layer, the adapters are still dense models encompassed with the linear layers in the attention module.
Others \citep{huang2023lorahub,yang2024moral, feng2024mixture,chen2024llava, wu2023mole} propose LoRA fusion based on the concept of Mixture-of-experts that enhance the model's ability for cross-domain tasks.
However, the methods mainly focus on fusing LoRA adapters to the FFN module or Q in the attention module. Our method could embed all kinds of LoRA adapters. By leveraging the full-mode MoE architecture, the LLM's capacity could be boosted with autonomous and timely LoRA switching, especially for solving composite tasks.

%% file: sections/limitations.tex
{\bf LoRA adapter update.} Although the Gating network within \method\ module is trained separately among the adapters, it is necessary to retrain or fine-tune the Gating network if some LoRA adapters are updated.
The Gating network is trained using the hidden state as inputs, which are influenced by LoRA adapters in previous layers. Testing revealed that directly replacing some LoRA adapters with improved versions did not enhance performance on our test set. However, after retraining the \method\ modules, the LLM equipped with \method\ exhibited performance improvements. Technically, this issue may be related to the domain shift problem, where the Gating network is applied to another operational field the distribution shift. Employing statistical methods such as \citep{10.1145/3368089.3409696, krishnan2020improving} may help calibrate the output of the Gating network to produce more accurate logits and results.

{\bf Knowledge fusion tasks.} 
Composite tasks, which involve a broad range of tasks, represent one type of complexity in terms of the scope of tasks.
More challenging are tasks that require knowledge fusion across domains.
To assess the capability of \method\ in knowledge fusion task, we construct a mathematics task by translating problems from GSM8K into a foreign language (e.g., Italian), so that the LLM with \method\ must solve these foreign language GSM8K problems by leveraging knowledge from both GSM8k LoRA (trained on problems in English) and the foreign language LoRA (trained for Italian to English translation). Although \method\ successfully fuses the two LoRA adapters to address the math problems in a foreign language, it does not show superior performance compared to LLM equipped only with the GSM8K LoRA. We hypothesize that the base LLM's existing proficiency in the selected foreign language may render the additional adapter unnecessary. Future efforts could focus on constructing more suitably complex tasks where the required cross-domain knowledge is not already pre-trained into the base LLM.

{\bf MoE efficiency.} 
Sparsely-gated MoE \citep{shazeer2017outrageously} offers computational efficiency advantages over dense MoE. However, the naive implementation of MoE forward (\textit{loop-original}), such as the SparseMoE in Mixtral \citep{jiang2024mixtral,hf_transformers}, still encounters efficiency issues when the number of experts increases.
In our evaluations, the runtime for inference can be up to $\mathbf{10 \times}$ longer than that of \textit{single-lora} when embedding 28 LoRA adapters into one LLM. With our proposed forward acceleration techniques \textit{bmm-torch} and \textit{bmm-triton}, we achieve a speedup of $\sim\!\mathbf{4 \times}$ compared to the \textit{loop-original}, though this still falls short of the ideal upper bound (\textit{single-lora}). Technically, it is extremely difficult to increase the inference speeds for \method\ when the number of embedded LoRA adapters increases.
Future work could explore developing new operators in triton or CUDA to continuously enhance MoE acceleration in terms of memory efficiency.

%% file: sections/conclusion.tex
This paper presents a framework \method\ that achieves scalable multi-task LoRA embedding within LLMs, enhancing the existing LLMs with a full-mode MoE architecture with forward acceleration strategies. 
LLMs equipped with \method\ enhance the ability to autonomously select the most pertinent LoRA adapters to generate appropriate responses. Moreover, its capability for timely LoRA switching leads to superior performance, particularly in sequentially solving composite tasks.
Future work could explore the transformative potential of \method\ in multifaceted problem-solving scenarios, and inference efficiency by designing more efficient GPU kernel operators.

%% file: sections/appendix.tex
\subsection{Top-k strategy}\label{appendix:topk}

\textbf{Top-$1$ strategy:} When the Gating network is configured to select the LoRA adapter with the maximum logit, the forward process of \method\ as detailed in Equation \ref{eq:forward} simplifies to the classical LoRA forward $h = W_{\mathrm{base}}x + B_iA_ix$. Thus, the weight $w_i$ calculated by the Gating network $G_i$ only contributes to the LoRA selection, but does not influence the token generation process, resulting in it being irrelevant to the loss $\mathcal{L}_{\mathrm{lm}}$. 
Thus, training of Gating networks under the top-$1$ strategy could utilize the following truncated loss function $\mathcal{L}_{\mathrm{top-}1}$:
\begin{equation}
\mathcal{L}_{\mathrm{top-}1}=\arg\max_{\theta}\sum_{i=1}^L\sum_{j=1}^B\sum_{k=1}^ml_{k,j}(h)
\end{equation}

\textbf{Top-$k$ strategy:}
With the top-$k$ strategy set in the Gating network, \method\ computes the normalized weights $w_i$ for the $k$ selected LoRA adapters.  These weights participate in the computation of the losses of both $\mathcal{L}_{\mathrm{lm}}$ and $\mathcal{L}_{\mathrm{gate}}$ as specified in Equation \ref{eq:forward}. Thus, the parameter updates for the Gating network derive from the losses associated with both LoRA classification and autoregressive token generation.
Notice that the LoRA classification loss is only influenced by the LoRA adapter with the highest logit, whereas the backpropagation from the token loss affects the parameters in the Gating network responsible for all $k$ selected LoRA adapters. Although these remaining $k-1$ adapters lack direct supervision from the LoRA classification loss, the token generation loss contributes to enhanced robustness and the capacity for LoRA switching during generation.

\subsection{Details on the tailored Triton kernel for efficient MeteoRA forward}\label{appendix: triton_kernel}

To address the memory copying problem caused by PyTorch indexing, we fuse the two $bmm$ operations inside a GPU kernel function implemented by Triton, which dynamically indexes the right pair of LoRA matrix $(A_i, B_i)$ and load them from HBM to SRAM in each parallelized thread. Therefore, there is no need to explicitly allocate $b\!\times\! s\!\times\! k$ pairs of $(A_i, B_i)$ over the original $n$ ones.

Another challenge is that Triton constraints all the dimensions for the matrix operators should be no less than $16$, however, under the MeteoRA settings, this requirement can never be satisfied since the first operator $\boldsymbol{x}$ is a vector, and also, the LoRA rank size may be less than $16$ easily (e.g., in all our experiments, we fix $r = 8$). Therefore, it is not that trivial to implement such a kernel, unless using the simple \textit{masking} strategy to meet the requirements with over $15\!\times$ waste of I/O.

\begin{algorithm}
\caption{Pseudo Code for BMM-Triton Kernel Function}\label{alg:bmm_triton}
\begin{algorithmic}[1]
\State Prepare blockized $X, A'$ with their masks $M_1, M_2$ before launching the kernel
\State Load $X$, $I$, $M_1$, $M_2$ from HBM to SRAM \Comment{$I$ is the candidate LoRA index set}
\State Load $A'$, $B$ indexed by $I$
\State $oA' = X \times A'$
\State $oA'' \gets \big(\big( oA' \odot M_1 \big) \times M_2\big)$
\State $oB' \gets oA'' \times B$
\State $O \gets \text{colsum}[oB']$ \Comment{Compute column-wise sum}
\State Store $O$ back from SRAM to HBM
\end{algorithmic}
\end{algorithm}

To both obey the dimension constraint and avoid too much waste, for the first $bmm$ of $\boldsymbol{x} \in \mathbb{R}^{1\times d}$ and $A \in \mathbb{R}^{d\times r}$, we use a \textit{blocking} strategy to split the vector $\boldsymbol{x}$ along the hidden size dimension by $m$ blocks, where $m >= 16$. In such case, the first operator becomes a matrix $X$ with shape $(m, \frac{d}{m})$, and also we have to split $A$ along the first dimension to become a more square matrix $A'$ with shape $(\frac{d}{m}, r\!\times\! m)$. Notice that now the output of first $bmm$: $oA' = X \times A'$ with shape $(m, r\!\times\! m)$ has a relationship with the original one $oA = x \times A$ with shape $(1, r)$ as follows:

\begin{equation}
\begin{aligned}
    oA'' &= \big(\big( oA' \odot M_1 \big) \times M_2\big)\\
    oA &= \text{colsum}[oA''] \label{eq:bmm-triton-loraA}
\end{aligned}
\end{equation}

where $M_1$ and $M_2$ are two trivial $01$ mask matrixs, each sized $(m, r\!\times\!m)$ and $(r\!\times\! m, r)$ respectively. So we can transform back to the right results by three additional negligible dot-product with $M_1$, matrix-product with $M_2$, and $\text{colsum}$ operations for the first $bmm$. For the second one, instead of directly using the right result $oA$, we can delay the $\text{colsum}$ operation until we finish the second $bmm$, i.e. we use $oA''$ with shape $(m, r)$ and $B$ with shape $(r, h)$ to do matrix-product operations to get the temporary result $oB'$ with shape $(m, h)$, then apply $\text{colsum}$ to get the final LoRA output $O$ with shape $(1, h)$. Notably, on one hand, we can avoid one more \textit{blocking} operation for $oA$ since $oA''$ already meets the dimension constraint, on the other hand, if $r < 16$, we can just simply utilize \textit{masking} strategy since it is the inner dimension and small enough.

Overall, for the Triton kernel function, we offer the pseudo code as shown in Algorithm~\ref{alg:bmm_triton}.

\newpage

\vspace{-\baselineskip}

\subsection{Information about 28 tasks}\label{appendix:28_task_details}

Table \ref{tab:28_tasks_details} shows the detailed information of the 28 selected tasks in the Section \ref{sec:experiment}. The name in parentheses is the abbreviation of the corresponding task.
We use the original training sets from these tasks to fine-tune the LoRA adapters and Gating networks in \method\ modules. 
To achieve a balanced fine-tuning across the diverse task spectrum and ensure efficient training, we construct a balanced dataset by randomly sampling 1,000 samples from each task. This balanced dataset is then divided into a training set with 25,200 samples (i.e., 900 samples for each task) and a validating set with 2,800 samples (i.e., 100 samples for each task) for fine-tuning. 
In terms of the evaluation part, the performances are evaluated on each task's original test set.

\begin{table}[h]
\centering
\caption{Details about the 28 selected tasks.}
\label{tab:28_tasks_details}
\renewcommand{\arraystretch}{1.4}
\resizebox{\columnwidth}{!}{
\begin{tabular}{cp{3.5cm}p{7cm}p{2.8cm}}
\toprule
Task Name & Keywords & Description & Evaluation Metrics \\ 
\midrule
abstract\_narrative\_understanding (AbsNarr) &
  narrative understanding, multiple choice &
  Given a narrative, choose the most related proverb. &
  Accuracy \\
alpaca (ALPACA) &
  instruction-tuning &
  Write appropriate answers according to instructions. &
  BLEU, ROUGE \\
cnn\_dailymail (CNNDM) &
  summarization &
  Given news articles, write the summarization. &
  ROUGE \\
contextual\_parametric\_knowledge\_conflicts (ConParaKC) &
  contextual question-answering, multiple choice &
  Answer questions given the contextual information. &
  Accuracy \\
cs\_algorithms (CSAlg) &
  algorithms, numerical response &
  Solve two common computer-science tasks. &
  Accuracy \\
disfl\_qa (DisflQA) &
  contextual question-answering, reading comprehension &
  Pick the correct answer span from the context given the disfluent question. &
  Accuracy \\
elementary\_math\_qa (ElemMath) &
  mathematics &
  Answer multiple choice mathematical word problems. &
  Accuracy \\
epistemic\_reasoning (EpiReason) &
  logical reasoning, multiple choice &
  Determine whether one sentence entails the next. &
  Accuracy \\
formal\_fallacies\_syllogisms\_negation (FormFall) &
  logical reasoning, multiple choice, &
  Distinguish deductively valid arguments from formal fallacies. &
  Accuracy \\
goal\_step\_wikihow (GSWiki) &
  causal reasoning, multiple choice &
  Perform one of three subtasks: step inference, goal inference, or step ordering. &
  Accuracy \\
gsm8k (GSM8K) &
  mathematics &
  Solve the grade school math word problems. &
  Accuracy \\
language\_identification (LangID) &
  multilingual, multiple choice &
  Given a sentence, select the correct language. &
  Accuracy \\
linguistics\_puzzles (LingPuzz) &
  logical reasoning, linguistics &
  Solve Rosetta Stone-style linguistics puzzles. &
  BLEU, ROUGE \\
logical\_deduction (LogDeduc) &
  logical reasoning, multiple choice &
  Deduce the order of a sequence of objects. &
  Accuracy \\
news\_commentary\_de (NewsDE) &
  multilingual, translation &
  Translate German sentences into English. &
  BLEU \\
news\_commentary\_es (NewsES) &
  multilingual, translation &
  Translate Spanish sentences into English. &
  BLEU \\
news\_commentary\_it (NewsIT) &
  multilingual, translation &
  Translate Italian sentences into English. &
  BLEU \\
object\_counting (ObjCount) &
  logical reasoning &
  Questions that involve enumerating objects and asking the model to count them. &
  Accuracy \\
paragraph\_segmentation (ParaSeg) &
  segmentation, multilingual &
  Identify the sentences that end a paragraph in a document. &
  Accuracy \\
play\_dialog\_same\_or\_different (PlayDiag) &
  reading comprehension, multiple choice &
  Determine if nearby lines in a Shakespeare play were spoken by the same individual. &
  Accuracy \\
question\_selection (QuestSel) &
  reading comprehension, multiple choice &
  Given an answer along with its context, select the most appropriate question which has the given answer as its answer. &
  Accuracy \\
reasoning\_about\_colored\_objects (ColorReason) &
  reading comprehension, logical reasoning, multiple choice &
  Answer extremely simple questions about the colors of objects on a surface. &
  Accuracy \\
strategyqa (StratQA) &
  logical reasoning, context-free question answering &
  Answer questions in which the required reasoning steps are implicit in the question. &
  BLEU, ROUGE, Accuracy \\
topical\_chat (TopChat) &
  free response &
  Open-domain response generation. &
  BLEU, ROUGE \\
tracking\_shuffled\_objects (TrackObj) &
  logical reasoning, multiple choice &
  Determine the final positions given initial positions and a description of a sequence of swaps. &
  Accuracy \\
unit\_conversion (UnitConv) &
  contextual question-answering, mathematics, multiple choice &
  Perform various tasks relating to units, including identification and conversion. &
  Accuracy \\
vitaminc\_fact\_verification (VitaFact) &
  truthfulness, reading comprehension, multiple choice &
  Identify whether a claim is True or False based on the given context. &
  Accuracy \\
winowhy (WinoWhy) &
  causal reasoning, multiple choice &
  Evaluate the reasoning in answering Winograd Schema Challenge questions. &
  Accuracy \\ 
\bottomrule
\end{tabular}}
\end{table}

\newpage

\vspace{-\baselineskip}

\subsection{Experimental results of 28 tasks}\label{apppendix:28_task_results}

Table \ref{tab:28_tasks_result_llama2_choice}, Table \ref{tab:28_tasks_result_llama3_choice}, Table \ref{tab:28_tasks_result_llama2_non_choice} and Table \ref{tab:28_tasks_result_llama3_non_choice} show the detailed evaluation results of different models on the 28 selected tasks. When drawing Figure \ref{fig:28_tasks_radar_graph}, for tasks we use BLEU and ROUGE as metrics, we selected BLEU for \textit{news\_commentary\_de}, \textit{news\_commentary\_es}, and \textit{news\_commentary\_it}, while opting for ROUGE-L for the remaining tasks.

\vspace{-0.8\baselineskip}

\begin{table}[H]
\centering
\caption{Experimental results for tasks using accuracy as metric (LlaMA2-13B base model).}
\label{tab:28_tasks_result_llama2_choice}
\resizebox{\columnwidth}{!}{
\begin{tabular}{ccccccccccccc}
\toprule
Task Name & \color{gray}{PEFT (reference)} & LoRA-F & LoRA-B & Avg LoRA & TIES & DARE & Arrow & \multicolumn{1}{c|}{LoraHub} & MeteoRA (T1-1k) & MeteoRA (T2-1k) & MeteoRA(T2-100) & MeteoRA(T2-5) \\
\midrule
AbsNarr & \multicolumn{1}{c|}{\color{gray}{0.863}} & 0.758 & 0.720 & 0.562 & 0.340 & 0.190 & 0.788 & \multicolumn{1}{c|}{0.278} & 0.858 & 0.860 & 0.860 & \textbf{0.868} \\
ConParaKC & \multicolumn{1}{c|}{\color{gray}{0.999}} & \textbf{0.999} & 0.994 & 0.424 & 0.579 & 0.554 & 0.836 & \multicolumn{1}{c|}{0.514} & \textbf{0.999} & \textbf{0.999} & \textbf{0.999} & 0.998 \\
CSAlg & \multicolumn{1}{c|}{\color{gray}{0.841}} & \textbf{0.848} & 0.818 & 0.333 & 0.504 & 0.572 & 0.712 & \multicolumn{1}{c|}{0.515} & 0.841 & 0.818 & 0.826 & 0.826 \\
DisflQA & \multicolumn{1}{c|}{\color{gray}{0.690}} & 0.670 & 0.573 & 0.306 & 0.356 & 0.307 & 0.506 & \multicolumn{1}{c|}{0.236} & 0.679 & \textbf{0.684} & 0.683 & 0.661 \\
ElemMath & \multicolumn{1}{c|}{\color{gray}{0.801}} & 0.671 & 0.375 & 0.707 & 0.249 & 0.212 & 0.369 & \multicolumn{1}{c|}{0.364} & \textbf{0.794} & 0.725 & 0.771 & 0.718 \\
EpiReason & \multicolumn{1}{c|}{\color{gray}{1.000}} & \textbf{1.000} & 0.995 & 0.367 & 0.390 & 0.367 & 0.685 & \multicolumn{1}{c|}{0.233} & \textbf{1.000} & 0.998 & \textbf{1.000} & \textbf{1.000} \\
FormFall & \multicolumn{1}{c|}{\color{gray}{0.999}} & 0.921 & 0.565 & 0.510 & 0.510 & 0.510 & 0.961 & \multicolumn{1}{c|}{0.299} & 0.999 & 0.996 & \textbf{1.000} & 0.999 \\
GSWiki & \multicolumn{1}{c|}{\color{gray}{0.906}} & 0.877 & 0.842 & 0.639 & 0.646 & 0.591 & 0.839 & \multicolumn{1}{c|}{0.260} & \textbf{0.887} & 0.872 & 0.879 & 0.881 \\
GSM8K & \multicolumn{1}{c|}{\color{gray}{0.458}} & 0.428 & 0.338 & 0.062 & 0.058 & 0.052 & 0.252 & \multicolumn{1}{c|}{0.155} & 0.420 & \textbf{0.439} & 0.427 & 0.397 \\
LangID & \multicolumn{1}{c|}{\color{gray}{0.874}} & 0.728 & 0.542 & 0.235 & 0.403 & 0.283 & 0.455 & \multicolumn{1}{c|}{0.253} & \textbf{0.872} & 0.854 & 0.869 & 0.848 \\
LogDeduc & \multicolumn{1}{c|}{\color{gray}{0.720}} & 0.653 & 0.680 & 0.330 & 0.360 & 0.323 & 0.587 & \multicolumn{1}{c|}{0.473} & 0.713 & 0.717 & 0.720 & \textbf{0.723} \\
ObjCount & \multicolumn{1}{c|}{\color{gray}{0.740}} & 0.690 & 0.725 & 0.330 & 0.285 & 0.245 & 0.290 & \multicolumn{1}{c|}{0.180} & 0.735 & 0.725 & \textbf{0.740} & 0.720 \\
ParaSeg & \multicolumn{1}{c|}{\color{gray}{0.214}} & 0.274 & 0.214 & 0.047 & 0.050 & 0.036 & 0.178 & \multicolumn{1}{c|}{0.015} & 0.195 & 0.182 & \textbf{0.297} & 0.295 \\
PlayDiag & \multicolumn{1}{c|}{\color{gray}{0.649}} & 0.649 & \textbf{0.650} & 0.649 & 0.649 & 0.649 & 0.649 & \multicolumn{1}{c|}{0.265} & 0.649 & 0.649 & 0.649 & 0.649 \\
QuestSel & \multicolumn{1}{c|}{\color{gray}{0.927}} & 0.801 & 0.794 & 0.509 & 0.617 & 0.506 & \textbf{0.937} & \multicolumn{1}{c|}{0.291} & \textbf{0.937} & 0.934 & 0.924 & 0.934 \\
ColorReason & \multicolumn{1}{c|}{\color{gray}{0.950}} & \textbf{0.950} & \textbf{0.950} & 0.400 & 0.400 & 0.393 & 0.660 & \multicolumn{1}{c|}{0.515} & 0.930 & 0.940 & 0.935 & 0.810 \\
StratQA & \multicolumn{1}{c|}{\color{gray}{0.731}} & 0.729 & 0.722 & 0.367 & 0.606 & 0.558 & 0.707 & \multicolumn{1}{c|}{0.573} & \textbf{0.742} & 0.722 & 0.718 & 0.722 \\
TrackObj & \multicolumn{1}{c|}{\color{gray}{0.188}} & 0.181 & 0.188 & 0.191 & 0.101 & 0.103 & 0.181 & \multicolumn{1}{c|}{0.125} & 0.173 & 0.192 & 0.185 & \textbf{0.195} \\
UnitConv & \multicolumn{1}{c|}{\color{gray}{0.755}} & \textbf{0.779} & 0.707 & 0.358 & 0.370 & 0.274 & 0.534 & \multicolumn{1}{c|}{0.308} & 0.727 & 0.735 & 0.729 & 0.604 \\
VitaFact & \multicolumn{1}{c|}{\color{gray}{0.899}} & \textbf{0.908} & 0.812 & 0.171 & 0.640 & 0.200 & 0.817 & \multicolumn{1}{c|}{0.245} & 0.897 & 0.897 & 0.897 & 0.893 \\
WinoWhy & \multicolumn{1}{c|}{\color{gray}{0.802}} & 0.797 & 0.767 & 0.002 & 0.028 & 0.038 & 0.005 & \multicolumn{1}{c|}{0.344} & 0.797 & 0.767 & \textbf{0.801} & 0.795 \\
\noalign{\smallskip}
\hline
\noalign{\smallskip}
Average & \multicolumn{1}{c|}{\color{gray}{0.762}} & 0.729 & 0.665 & 0.357 & 0.388 & 0.332 & 0.569 & \multicolumn{1}{c|}{0.307} & 0.754 & 0.748 & \textbf{0.758} & 0.740 \\
\bottomrule
\end{tabular}}
\end{table}

\vspace{-0.8\baselineskip}

\begin{table}[H]
\centering
\caption{Experimental results for tasks using accuracy as metric (LlaMA3-8B base model).}
\label{tab:28_tasks_result_llama3_choice}
\resizebox{\columnwidth}{!}{
\begin{tabular}{ccccccccccccc}
\toprule
Task Name                          & \color{gray}{PEFT (reference)}                       & LORA-F & LORA-B & Avg LoRA & TIES & DARE & Arrow & LoraHub & MeteoRA (T1-1k) & MeteoRA (T2-1k) & MeteoRA(T2-100) & MeteoRA(T2-5) \\
\midrule
AbsNarr & \multicolumn{1}{c|}{\color{gray}{0.803}} & \textbf{0.793} & 0.790 & 0.413 & 0.425 & 0.335 & 0.772 & \multicolumn{1}{c|}{0.075} & 0.787 & 0.787 & 0.775 & 0.768 \\
ConParaKC & \multicolumn{1}{c|}{\color{gray}{0.999}} & \textbf{0.999} & \textbf{0.999} & 0.514 & 0.594 & 0.492 & 0.997 & \multicolumn{1}{c|}{0.219} & \textbf{0.999} & \textbf{0.999} & 0.976 & 0.992 \\
CSAlg                     & \multicolumn{1}{c|}{\color{gray}{0.841}} & 0.841 & 0.841 & 0.705 & 0.686 & 0.663 & 0.780 & \multicolumn{1}{c|}{0.602} & \textbf{0.845} & 0.826 & 0.826 & 0.830 \\
DisflQA                          & \multicolumn{1}{c|}{\color{gray}{0.703}} & 0.680 & 0.605 & 0.374 & 0.396 & 0.377 & 0.504 & \multicolumn{1}{c|}{0.197} & \textbf{0.706} & 0.703 & 0.686 & 0.628 \\
ElemMath               & \multicolumn{1}{c|}{\color{gray}{0.780}} &  \textbf{0.777} & 0.606 & 0.273 & 0.308 & 0.245 & 0.645 & \multicolumn{1}{c|}{0.106} & 0.776 & 0.773 & 0.751 & 0.725 \\
EpiReason               & \multicolumn{1}{c|}{\color{gray}{1.000}} &  0.996 & \textbf{1.000} & 0.430 & 0.450 & 0.425 & 0.600 & \multicolumn{1}{c|}{0.170} & \textbf{1.000} & \textbf{1.000} & \textbf{1.000} & \textbf{1.000}  \\
FormFall      & \multicolumn{1}{c|}{\color{gray}{0.989}} & 0.970 & 0.628 & 0.528 & 0.519 & 0.520 & 0.836 & \multicolumn{1}{c|}{0.190} & \textbf{0.987} & \textbf{0.987} & 0.981 & 0.977\\
GSWiki                & \multicolumn{1}{c|}{\color{gray}{0.935}} & 0.921 & 0.923 & 0.627 & 0.608 & 0.574 & 0.835 & \multicolumn{1}{c|}{0.307} & \textbf{0.932} & 0.928 & 0.904 & 0.896 \\
GSM8K                              & \multicolumn{1}{c|}{\color{gray}{0.591}} & 0.566 & 0.548 & 0.080 & 0.086 & 0.108 & 0.172 & \multicolumn{1}{c|}{0.050} & 0.555 & \textbf{0.559} & 0.511 & 0.491 \\
LangID           & \multicolumn{1}{c|}{\color{gray}{0.782}} & 0.749 & 0.649 & 0.404 & 0.412 & 0.383 & 0.625 & \multicolumn{1}{c|}{0.192} & \textbf{0.779} & 0.775 & 0.759 & 0.744   \\
LogDeduc                 & \multicolumn{1}{c|}{\color{gray}{0.760}} & 0.707 & 0.707 & 0.403 & 0.423 & 0.383 & 0.627 & \multicolumn{1}{c|}{0.367} & 0.757 & 0.753 & 0.747 & \textbf{0.770}   \\
ObjCount                   & \multicolumn{1}{c|}{\color{gray}{0.880}} & 0.555 & 0.865 & 0.060 & 0.080 & 0.130 & 0.005 & \multicolumn{1}{c|}{0.230} & \textbf{0.875} & 0.850 & 0.785 & 0.750   \\
ParaSeg            & \multicolumn{1}{c|}{\color{gray}{0.296}} & 0.261 & 0.244 & 0.044 & 0.050 & 0.045 & 0.187 & \multicolumn{1}{c|}{0.000} & \textbf{0.295} & 0.252 & 0.235 & 0.234   \\
PlayDiag  & \multicolumn{1}{c|}{\color{gray}{0.649}} & 0.632 & 0.649 & 0.647 & 0.650 & 0.644 & \textbf{0.656} & \multicolumn{1}{c|}{0.092} & 0.649 & 0.649 & 0.580 & 0.581   \\
QuestSel                & \multicolumn{1}{c|}{\color{gray}{0.936}} & 0.911 & 0.930 & 0.544 & 0.506 & 0.472 & 0.845 & \multicolumn{1}{c|}{0.247} & 0.927 & \textbf{0.940} & 0.892 & 0.892   \\
ColorReason & \multicolumn{1}{c|}{\color{gray}{0.958}} & 0.945 & \textbf{0.965} & 0.565 & 0.595 & 0.530 & 0.793 & \multicolumn{1}{c|}{0.238} & 0.960 & 0.983 & 0.915 & 0.905   \\
StratQA                         & \multicolumn{1}{c|}{\color{gray}{0.716}} & 0.707 & \textbf{0.718} & 0.600 & 0.611 & 0.538 & 0.681 & \multicolumn{1}{c|}{0.503} & 0.659 & 0.670 & 0.648 & 0.611   \\
TrackObj        & \multicolumn{1}{c|}{\color{gray}{0.995}} & 0.588 & 0.664 & 0.147 & 0.195 & 0.136 & 0.804 & \multicolumn{1}{c|}{0.171} & 0.993 & \textbf{0.996} & 0.985 & 0.985  \\
UnitConv                   & \multicolumn{1}{c|}{\color{gray}{0.822}} & 0.814 & 0.780 & 0.485 & 0.491 & 0.410 & 0.647 & \multicolumn{1}{c|}{0.463} & \textbf{0.820} & 0.819 & 0.802 & 0.786   \\
VitaFact       & \multicolumn{1}{c|}{\color{gray}{0.908}} &  0.903 & 0.839 & 0.607 & 0.655 & 0.541 & 0.822 & \multicolumn{1}{c|}{0.311} & \textbf{0.907} & \textbf{0.907} & 0.902 & 0.890  \\
WinoWhy                            & \multicolumn{1}{c|}{\color{gray}{0.816}} & 0.797 & 0.802 & 0.524 & 0.516 & 0.526 & 0.750 & \multicolumn{1}{c|}{0.203} & 0.818 & \textbf{0.827} & 0.788 & 0.788   \\ 
\noalign{\smallskip}
\hline
\noalign{\smallskip}
Average                            & \multicolumn{1}{c|}{\color{gray}{0.817}} & 0.767 & 0.750 & 0.427 & 0.441 & 0.404 & 0.647 & \multicolumn{1}{c|}{0.235} & \textbf{0.811} & 0.806 & 0.783 & 0.773   \\
\bottomrule
\end{tabular}}
\end{table}

\vspace{-0.8\baselineskip}

\newpage
\begin{longtable}{cccccc}
\caption{Experimental results for tasks using BLEU and ROUGE as metrics (LlaMA2-13B base model).} \label{tab:28_tasks_result_llama2_non_choice} \\
\toprule
Task Name & Model & BLEU  & ROUGE-1 & ROUGE-2 & ROUGE-L \\
\midrule
\endfirsthead

\toprule
Task Name & Model & BLEU  & ROUGE-1 & ROUGE-2 & ROUGE-L \\
\midrule
\endhead

\multirow{12}{*}{ALPACA}               & \color{gray}{PEFT (reference)}            & \color{gray}{16.03} & \color{gray}{0.363}   & \color{gray}{0.176}   & \color{gray}{0.340}   \\ \noalign{\smallskip} \cline{2-6} \noalign{\smallskip}
                                      & LoRA-F         & 23.96 & 0.302   & 0.140   & 0.283   \\
                                      & LoRA-B         & 11.72 & 0.341   & 0.157   & 0.317   \\
                                      & Avg LoRA       & 41.88 & 0.195   & 0.084   & 0.164   \\
                                      & TIES           & \textbf{80.34} & 0.209   & 0.092   & 0.175   \\
                                      & DARE           & 78.25 & 0.228   & 0.101   & 0.193   \\
                                      & Arrow          & 24.62 & 0.271   & 0.128   & 0.230   \\
                                      & LoraHub        & 0.00  & 0.240   & 0.117   & 0.206   \\
                                      \noalign{\smallskip}
                                      \cline{2-6}
                                      \noalign{\smallskip}
                                      & MeteoRA (T1-1k) & 28.83 & \textbf{0.350}   & \textbf{0.166}   & \textbf{0.329}   \\
                                      & MeteoRA (T2-1k) & 24.12 & 0.349   & 0.162   & 0.327   \\
                                      & MeteoRA (T2-100) & 39.09 & 0.332 & 0.160 & 0.281   \\
                                      & MeteoRA (T2-5) & 12.49 & 0.306 & 0.140 & 0.256   \\
                                      \noalign{\smallskip}
                                      \hline
                                      \noalign{\smallskip}
\multirow{12}{*}{CNNDM}       & \color{gray}{PEFT (reference)}            & \color{gray}{7.50}  & \color{gray}{0.228}   & \color{gray}{0.067}   & \color{gray}{0.214}   \\ \noalign{\smallskip} \cline{2-6} \noalign{\smallskip} 
                                      & LoRA-F         & 15.69 & 0.241   & \textbf{0.076}   & \textbf{0.227}   \\
                                      & LoRA-B         & 15.65 & 0.228   & 0.067   & 0.214   \\
                                      & Avg LoRA       & 13.08 & 0.144   & 0.032   & 0.104   \\
                                      & TIES           & 13.08 & 0.147   & 0.032   & 0.104   \\
                                      & DARE           & 13.08 & 0.126   & 0.031   & 0.081   \\
                                      & Arrow          & \textbf{17.42} & 0.173   & 0.043   & 0.122   \\
                                      & LoraHub        & 4.77  & 0.141   & 0.030   & 0.104   \\
                                      \noalign{\smallskip}
                                      \cline{2-6}
                                      \noalign{\smallskip}
                                      & MeteoRA (T1-1k) & 7.50  & 0.229   & 0.069   & 0.216   \\
                                      & MeteoRA (T2-1k) & 5.57  & 0.230   & 0.070   & 0.217   \\
                                      & MeteoRA (T2-100) & 7.32 & 0.251   & 0.070   & 0.196   \\
                                      & MeteoRA (T2-5) & 7.77 & \textbf{0.254}   & 0.073   & 0.199   \\
                                      \noalign{\smallskip}
                                      \hline
                                      \noalign{\smallskip}
\multirow{12}{*}{LingPuzz} & \color{gray}{PEFT (reference)}            & \color{gray}{46.17} & \color{gray}{0.716}   & \color{gray}{0.479}   & \color{gray}{0.659}   \\ \noalign{\smallskip} \cline{2-6} \noalign{\smallskip}
                                      & LoRA-F         & 62.23 & 0.649   & 0.365   & 0.582   \\
                                      & LoRA-B         & 54.91 & 0.608   & 0.324   & 0.541   \\
                                      & Avg Lora       & 36.72 & 0.531   & 0.233   & 0.441   \\
                                      & TIES           & 49.14 & 0.405   & 0.117   & 0.308   \\
                                      & DARE           & \textbf{68.87} & 0.379   & 0.102   & 0.285   \\
                                      & Arrow          & 56.23 & 0.643   & 0.365   & 0.562   \\
                                      & LoraHub        & 0.00  & 0.172   & 0.057   & 0.131   \\
                                      \noalign{\smallskip}
                                      \cline{2-6}
                                      \noalign{\smallskip}
                                      & MeteoRA (T1-1k) & 68.34 & 0.717   & 0.478   & \textbf{0.661}   \\
                                      & MeteoRA (T2-1k) & 46.17 & 0.713   & 0.476   & 0.655   \\
                                      & MeteoRA (T2-100) & 46.17 & \textbf{0.718}   & \textbf{0.480}   & 0.646   \\
                                      & MeteoRA (T2-5) & 57.47 & 0.716   & 0.474   & 0.646   \\
                                      \noalign{\smallskip}
                                      \hline
                                      \noalign{\smallskip}
\multirow{12}{*}{NewsDE} & \color{gray}{PEFT (reference)}            & \color{gray}{78.25} & -       & -       & -       \\ \noalign{\smallskip} \cline{2-6} \noalign{\smallskip}
                                      & LoRA-F         & 78.25 & -       & -       & -       \\
                                      & LoRA-B         & 78.25 & -       & -       & -       \\
                                      & Avg Lora       & 3.38  & -       & -       & -       \\
                                      & TIES           & \textbf{86.48} & -       & -       & -       \\
                                      & DARE           & \textbf{86.48} & -       & -       & -       \\
                                      & Arrow          & \textbf{86.48} & -       & -       & -       \\
                                      & LoraHub        & 50.09 & -       & -       & -       \\
                                      \noalign{\smallskip}
                                      \cline{2-6}
                                      \noalign{\smallskip}
                                      & MeteoRA (T1-1k) & \textbf{86.48} & -       & -       & -       \\
                                      & MeteoRA (T2-1k) & \textbf{86.48} & -       & -       & -       \\
                                      & MeteoRA (T2-100) & \textbf{86.48} & -       & -       & -       \\
                                      & MeteoRA (T2-5) & \textbf{86.48} & -       & -       & -       \\
                                      \noalign{\smallskip}
                                      \hline
                                      \noalign{\smallskip}
                                      \pagebreak
\multirow{12}{*}{NewsES} & \color{gray}{PEFT (reference)}            & \color{gray}{70.05} & -       & -       & -       \\ \noalign{\smallskip} \cline{2-6} \noalign{\smallskip} 
                                      & LoRA-F         & 57.03 & -       & -       & -       \\
                                      & LoRA-B         & \textbf{81.54} & -       & -       & -       \\
                                      & Avg Lora       & 2.86  & -       & -       & -       \\
                                      & TIES           & 70.05 & -       & -       & -       \\
                                      & DARE           & 46.27 & -       & -       & -       \\
                                      & Arrow          & \textbf{81.54} & -       & -       & -       \\
                                      & LoraHub        & 0.64 & -       & -       & -       \\
                                      \noalign{\smallskip}
                                      \cline{2-6}
                                      \noalign{\smallskip}
                                      & MeteoRA (T1-1k) & \textbf{81.54} & -       & -       & -       \\
                                      & MeteoRA (T2-1k) & 70.05 & -       & -       & -       \\
                                      & MeteoRA (T2-100) & 70.05 & -       & -       & -       \\
                                      & MeteoRA (T2-5) & 70.05 & -       & -       & -       \\
                                      \noalign{\smallskip}
                                      \hline
                                      \noalign{\smallskip}
\multirow{12}{*}{NewsIT} & \color{gray}{PEFT (reference)}            & \color{gray}{39.04} & -       & -       & -       \\ \noalign{\smallskip} \cline{2-6} \noalign{\smallskip} 
                                      & LoRA-F         & \textbf{54.90} & -       & -       & -       \\
                                      & LoRA-B         & 40.08 & -       & -       & -       \\\
                                      & Avg Lora       & 40.20  & -       & -       & -       \\
                                      & TIES           & 40.08 & -       & -       & -       \\
                                      & DARE           & 40.20 & -       & -       & -       \\
                                      & Arrow          & 36.92 & -       & -       & -       \\
                                      & LoraHub        & 40.08 & -       & -       & -       \\
                                      \noalign{\smallskip}
                                      \cline{2-6}
                                      \noalign{\smallskip}
                                      & MeteoRA (T1-1k) & 39.04 & -       & -       & -       \\
                                      & MeteoRA (T2-1k) & 39.04 & -       & -       & -       \\
                                      & MeteoRA (T2-100) & 39.04 & -       & -       & -       \\
                                      & MeteoRA (T2-1k) & 39.04 & -       & -       & -       \\
                                      \noalign{\smallskip}
                                      \hline
                                      \noalign{\smallskip}
\multirow{12}{*}{StratQA}           & \color{gray}{PEFT (reference)}            & \color{gray}{15.72} & \color{gray}{0.237}   & \color{gray}{0.064}   & \color{gray}{0.222}   \\ \noalign{\smallskip} \cline{2-6} \noalign{\smallskip} 
                                      & LoRA-F         & 9.71  & 0.247   & \textbf{0.076}   & \textbf{0.238}   \\
                                      & LoRA-B         & 11.90 & \textbf{0.249}   & 0.073   & 0.236   \\
                                      & Avg Lora       & 14.54 & 0.185   & 0.050   & 0.149   \\
                                      & TIES           & 16.62 & 0.112   & 0.024   & 0.088   \\
                                      & DARE           & 16.62 & 0.123   & 0.026   & 0.095   \\
                                      & Arrow          & 13.83 & 0.218   & 0.066   & 0.175   \\
                                      & LoraHub        & 11.50 & 0.171   & 0.038   & 0.128   \\
                                      \noalign{\smallskip}
                                      \cline{2-6}
                                      \noalign{\smallskip}
                                      & MeteoRA (T1-1k) & 8.74  & 0.235   & 0.065   & 0.221   \\
                                      & MeteoRA (T2-1k) & 10.03 & 0.240   & 0.068   & 0.226   \\
                                      & MeteoRA (T2-100) & 13.95 & 0.222  & 0.063   & 0.172   \\
                                      & MeteoRA (T2-5) & \textbf{20.69} & 0.228    & 0.067   & 0.174   \\
                                      \noalign{\smallskip}
                                      \hline
                                      \noalign{\smallskip}
\multirow{12}{*}{TopChat}        & \color{gray}{PEFT (reference)}            & \color{gray}{12.50} & \color{gray}{0.157}   & \color{gray}{0.027}   & \color{gray}{0.146}   \\ \noalign{\smallskip} \cline{2-6} \noalign{\smallskip}
                                      & LoRA-F         & \textbf{28.39} & \textbf{0.153}   & \textbf{0.025}   & \textbf{0.142}   \\
                                      & LoRA-B         & 9.78  & 0.143   & 0.021   & 0.134   \\
                                      & Avg Lora       & 1.21  & 0.099   & 0.010   & 0.060   \\
                                      & TIES           & 22.45 & 0.101   & 0.011   & 0.078   \\
                                      & DARE           & 22.48 & 0.103   & 0.011   & 0.065   \\
                                      & Arrow          & 11.16 & 0.099   & 0.013   & 0.080   \\
                                      & LoraHub        & 0.35  & 0.064   & 0.005   & 0.051   \\
                                      \noalign{\smallskip}
                                      \cline{2-6}
                                      \noalign{\smallskip}
                                      & MeteoRA (T1-1k) & 13.44 & 0.151   & \textbf{0.025}   & 0.141   \\
                                      & MeteoRA (T2-1k) & 12.35 & 0.149   & \textbf{0.025}   & 0.140   \\
                                      & MeteoRA (T2-100) & 13.44 & 0.132  & 0.023   & 0.108   \\
                                      & MeteoRA (T2-5) & 12.93 & 0.135 & 0.024 & 0.110   \\
\bottomrule
\end{longtable}

\vspace{-0.8\baselineskip}

\begin{longtable}{cccccc}
\caption{Experimental results for tasks using BLEU and ROUGE as metrics (LlaMA3-8B base model).} \label{tab:28_tasks_result_llama3_non_choice} \\
\toprule
Task Name & Model & BLEU  & ROUGE-1 & ROUGE-2 & ROUGE-L \\
\midrule
\endfirsthead

\toprule
Task Name & Model & BLEU  & ROUGE-1 & ROUGE-2 & ROUGE-L \\
\midrule
\endhead

\multirow{12}{*}{ALPACA}               & \color{gray}{PEFT (reference)}            & \color{gray}{24.72} & \color{gray}{0.376}   & \color{gray}{0.190}   & \color{gray}{0.353}   \\ \noalign{\smallskip} \cline{2-6} \noalign{\smallskip}
                                      & LoRA-F         & 31.47 & 0.284   & 0.123   & 0.267   \\
                                      & LoRA-B         & 29.27 & \textbf{0.358}   & \textbf{0.175}   & \textbf{0.335}   \\
                                      & Avg LoRA & 73.49 & 0.206 & 0.089 & 0.172\\
                                      & TIES & 73.49 & 0.214 & 0.092 & 0.181\\
                                      & DARE & 73.49 & 0.230 & 0.099 & 0.192\\
                                      & Arrow & 12.26 & 0.222 & 0.093 & 0.186\\
                                      & LoraHub & 0.00 & 0.176 & 0.068 & 0.151\\
                                      \noalign{\smallskip}
                                      \cline{2-6}
                                      \noalign{\smallskip}
                                      & MeteoRA (T1-1k) & 32.34 & \textbf{0.358}   & 0.170   & \textbf{0.335}   \\
                                      & MeteoRA (T2-1k) & 30.08 & 0.354   & 0.170   & 0.332   \\
                                      & MeteoRA (T2-100) & 31.19 & 0.317 & 0.147 & 0.266\\
                                      & MeteoRA (T2-5) & \textbf{80.34} & 0.249 & 0.103 & 0.204\\
                                      \noalign{\smallskip}
                                      \hline
                                      \noalign{\smallskip}
\multirow{12}{*}{CNNDM}       & \color{gray}{PEFT (reference)}            & \color{gray}{11.93} & \color{gray}{0.231}   & \color{gray}{0.069}   & \color{gray}{0.218}   \\ \noalign{\smallskip} \cline{2-6} \noalign{\smallskip}
                                      & LoRA-F         & 16.13 & \textbf{0.248}   & \textbf{0.080}   & \textbf{0.233}   \\
                                      & LoRA-B         & 13.27 & 0.233   & 0.070   & 0.218   \\
                                      & Avg LoRA & 21.07 & 0.168   & 0.039   & 0.121   \\
                                      & TIES & 18.07 &  0.154  & 0.037   & 0.109   \\
                                      & DARE & 4.67 & 0.137   & 0.032   & 0.096   \\
                                      & Arrow & 13.13 & 0.153   & 0.037   & 0.111   \\
                                      & LoraHub & 15.30 & 0.087   & 0.008   & 0.038   \\
                                      \noalign{\smallskip}
                                      \cline{2-6}
                                      \noalign{\smallskip}
                                      & MeteoRA (T1-1k) & 11.93 & 0.233   & 0.070   & 0.218   \\
                                      & MeteoRA (T2-1k) & 11.93 & 0.232   & 0.070   & 0.219   \\
                                      & MeteoRA (T2-100) & \textbf{21.11} & 0.205   & 0.054   & 0.146   \\
                                      & MeteoRA (T2-5) & 6.52 & 0.203   & 0.054   & 0.143   \\
                                      \noalign{\smallskip}
                                      \hline
                                      \noalign{\smallskip}
\multirow{12}{*}{LingPuzz} & \color{gray}{PEFT (reference)}            & \color{gray}{44.12} & \color{gray}{0.785} & \color{gray}{0.589}   & \color{gray}{0.734} \\ \noalign{\smallskip} \cline{2-6} \noalign{\smallskip}
                                      & LoRA-F         & 36.89 & 0.718   & 0.488   & 0.666   \\
                                      & LoRA-B         & 37.10 & \textbf{0.743}   & \textbf{0.519}   & \textbf{0.689}   \\
                                      & Avg LoRA & 28.87 & 0.421   & 0.134   & 0.331   \\
                                      & TIES & 34.17 & 0.432   & 0.134   & 0.339   \\
                                      & DARE & 56.23 & 0.357   & 0.113   & 0.281   \\
                                      & Arrow & \textbf{59.00} & 0.721   & 0.505   & 0.659   \\
                                      & LoraHub & 39.28 & 0.245   & 0.063   & 0.184   \\
                                      \noalign{\smallskip}
                                      \cline{2-6}
                                      \noalign{\smallskip}
                                      & MeteoRA (T1-1k) & 41.72 & 0.695   & 0.451   & 0.636   \\
                                      & MeteoRA (T2-1k) & 41.72 & 0.696   & 0.448   & 0.639   \\
                                      & MeteoRA (T2-100) & 50.81 & 0.666   & 0.408   & 0.588   \\
                                      & MeteoRA (T2-5) & 46.17 & 0.655   & 0.394   & 0.580   \\
                                      \noalign{\smallskip}
                                      \hline
                                      \noalign{\smallskip}
\multirow{12}{*}{NewsDE} & \color{gray}{PEFT (reference)}            & \color{gray}{97.65} & -       & -       & -       \\ \noalign{\smallskip} \cline{2-6} \noalign{\smallskip}
                                      & LoRA-F         & 78.25 & -       & -       & -       \\
                                      & LoRA-B         & 78.25 & -       & -       & -       \\
                                      & Avg LoRA & 63.56 & -   & -   & -   \\
                                      & TIES & 46.47 & -   & -   & -   \\
                                      & DARE & 36.60 & -   & -   & -   \\
                                      & Arrow & 37.36 & -   & -   & -   \\
                                      & LoraHub & 11.87 & -   & -   & -   \\
                                      \noalign{\smallskip}
                                      \cline{2-6}
                                      \noalign{\smallskip}
                                      & MeteoRA (T1-1k) & \textbf{86.48} & -       & -       & -       \\
                                      & MeteoRA (T2-1k) & \textbf{86.48} & -       & -       & -       \\
                                      & MeteoRA (T2-100) & 51.42 & -   & -   & -   \\
                                      & MeteoRA (T2-5) & \textbf{86.48} & -   & -   & -   \\
                                      \noalign{\smallskip}
                                      \hline
                                      \noalign{\smallskip}
                                      \pagebreak
\multirow{12}{*}{NewsES} & \color{gray}{PEFT (reference)}            & \color{gray}{81.54} & -       & -       & -       \\ \noalign{\smallskip} \cline{2-6} \noalign{\smallskip}
                                      & LoRA-F         & \textbf{81.54} & -       & -       & -       \\
                                      & LoRA-B         & \textbf{81.54} & -       & -       & -       \\
                                      & Avg LoRA & 31.18 & -   & -   & -   \\
                                      & TIES & 30.55 & -   & -   & -   \\
                                      & DARE & 17.61 & -   & -   & -   \\
                                      & Arrow & 31.82 & -   & -   & -   \\
                                      & LoraHub & 0.0 & -   & -   & -   \\
                                      \noalign{\smallskip}
                                      \cline{2-6}
                                      \noalign{\smallskip}
                                      & MeteoRA (T1-1k) & \textbf{81.54} & -       & -       & -       \\
                                      & MeteoRA (T2-1k) & \textbf{81.54} & -       & -       & -       \\
                                      & MeteoRA (T2-100) & \textbf{81.54} & -   & -   & -   \\
                                      & MeteoRA (T2-5) & 63.72 & -   & -   & -   \\
                                      \noalign{\smallskip}
                                      \hline
                                      \noalign{\smallskip}
\multirow{12}{*}{NewsIT} & \color{gray}{PEFT (reference)}            & \color{gray}{54.90} & -       & -       & -       \\ \noalign{\smallskip} \cline{2-6} \noalign{\smallskip}
                                      & LoRA-F         & \textbf{54.90} & -       & -       & -       \\
                                      & LoRA-B         & 38.54 & -       & -       & -       \\
                                      & Avg LoRA & 38.54 & -   & -   & -   \\
                                      & TIES & 37.48 & -   & -   & -   \\
                                      & DARE & 52.21 & -   & -   & -   \\
                                      & Arrow & 38.02 & -   & -   & -   \\
                                      & LoraHub & 0.0 & -   & -   & -   \\
                                      \noalign{\smallskip}
                                      \cline{2-6}
                                      \noalign{\smallskip}
                                      & MeteoRA (T1-1k) & \textbf{54.90} & -       & -       & -       \\
                                      & MeteoRA (T2-1k) & 51.83 & -       & -       & -       \\
                                      & MeteoRA (T2-100) & 35.22 & -   & -   & -   \\
                                      & MeteoRA (T2-5) & 36.78 & -   & -   & -   \\
                                      \noalign{\smallskip}
                                      \hline
                                      \noalign{\smallskip}
\multirow{12}{*}{StratQA}           & \color{gray}{PEFT (reference)}            & \color{gray}{10.58} & \color{gray}{0.249}   & \color{gray}{0.077}   & \color{gray}{0.236}   \\ \noalign{\smallskip} \cline{2-6} \noalign{\smallskip} 
                                      & LoRA-F         & 10.44 & 0.234   & 0.068   & 0.223   \\
                                      & LoRA-B         & 10.58 & 0.243   & 0.071   & 0.230   \\
                                      & Avg LoRA & \textbf{38.80} & 0.112   & 0.024   & 0.089   \\
                                      & TIES & 10.90 & 0.102   & 0.022   & 0.082   \\
                                      & DARE & 14.78 & 0.128   & 0.027   & 0.100   \\
                                      & Arrow & 12.19 & 0.206   & 0.057   & 0.165   \\
                                      & LoraHub & 14.35 & 0.147   & 0.033   & 0.116   \\
                                      \noalign{\smallskip}
                                      \cline{2-6}
                                      \noalign{\smallskip}
                                      & MeteoRA (T1-1k) &10.58 & \textbf{0.252}   & 0.076   & \textbf{0.239}   \\
                                      & MeteoRA (T2-1k) & 10.58 & 0.250   & \textbf{0.077}   & \textbf{0.239}   \\
                                      & MeteoRA (T2-100) & 20.56 & 0.228   & 0.065   & 0.174   \\
                                      & MeteoRA (T2-5) & 11.67 & 0.213   & 0.055   & 0.162   \\
                                      \noalign{\smallskip}
                                      \hline
                                      \noalign{\smallskip}
\multirow{12}{*}{TopChat}        & \color{gray}{PEFT (reference)}            & \color{gray}{39.50} & \color{gray}{0.151}   & \color{gray}{0.025}   & \color{gray}{0.141}   \\ \noalign{\smallskip} \cline{2-6} \noalign{\smallskip}
                                      & LoRA-F         & 33.82 & 0.150   & 0.024   & 0.140   \\
                                      & LoRA-B         & 19.22 & 0.139   & 0.019   & 0.131   \\
                                      & Avg LoRA & 23.59 & 0.094   & 0.012   & 0.078   \\
                                      & TIES & 26.13 & 0.092   & 0.011   & 0.077   \\
                                      & DARE & 38.31 & 0.086   & 0.008   & 0.066   \\
                                      & Arrow & 35.64 & 0.112   & 0.016   & 0.091   \\
                                      & LoraHub & 0.08 & 0.049   & 0.002   & 0.031   \\
                                      \noalign{\smallskip}
                                      \cline{2-6}
                                      \noalign{\smallskip}
                                      & MeteoRA (T1-1k) & \textbf{45.64} & \textbf{0.152}   & \textbf{0.026}   & \textbf{0.141}   \\
                                      & MeteoRA (T2-1k) & \textbf{45.64} & \textbf{0.152}   & 0.024   & \textbf{0.141}   \\
                                      & MeteoRA (T2-100) & 27.36 & 0.129   & 0.021   & 0.107   \\
                                      & MeteoRA (T2-5) & 40.86 & 0.130   & 0.018   & 0.109   \\
\bottomrule
\end{longtable}

\newpage
\subsection{\textit{Composite-n} evaluation results details}\label{sec:serial_result_details}

The tasks for constructing \textit{composite-n} are selected from the aforementioned set of 28 tasks to ensure the models familiarity and potential problem-solving capability. However, given the limited capability of the instruction following in the zero-shot setting, neither the \method\ models nor the models fine-tuned by LoRA achieve satisfactory results. Hence, we employ a 2-shot setting for evaluation on these \textit{composite-n} tasks.

The evaluation metrics used for \textit{composite-n} tasks are: average number of questions attempted, average number of multiple-choice questions answered correctly, and average BLEU, ROUGE scores for non-multiple-choice questions.

Notice that in the \textit{composite-n tasks}, when calculating the softmax values of the weights for the two LoRA adapters selected by the Gating network, we introduced a hyperparameter called \textit{temperature}. The value of \textit{temperature} needs to be increased as the number of sub-tasks grows. Specifically, we set the \textit{temperature} values to $15$, $20$, and $30$ for the three tasks, respectively.

Tables \ref{tab:serial_3_eval_results}, \ref{tab:serial_5_eval_results}, and \ref{tab:serial_10_eval_results} present the detailed evaluation results for the \textit{composite-3}, \textit{composite-5}, and \textit{composite-10} tasks, respectively. Several important clarifications are necessary for interpreting these results:
\begin{enumerate}
    \item The models are required to generate both the corresponding question number and its answer. Any mismatch between the question number and the answer is therefore considered incorrect.
    \item In the evaluation results, some BLEU scores are recorded as $0$. This occurs when the model generates mismatched question numbers and answers or provides extremely insufficient answers, resulting in an overall $0$ BLEU score.
    \item For the task \textit{strategyqa}, which involves answering with either 'yes' or 'no' and providing reasoning steps, the accuracy metric specifically measures the correctness of the 'yes' or 'no' response.
    \item The reported ROUGE scores refer to the F1-scores.
    \item Samples that the lengths exceed to 4,096 tokens are skipped in the evaluation process (we skip 13 samples in total).
\end{enumerate}

\begin{table}[ht]
\centering
\caption{The \textit{composite-3} evaluation results are presented in details with MeteoRA results on the left side and LoRA-B results on the right side of each metric column. A dash ('-') indicates that the corresponding metric was not applicable or included in the evaluation.
}
\label{tab:serial_3_eval_results}
\resizebox{\columnwidth}{!}{
\begin{tabular}{ccccccccccc}
\toprule
Sub-task Name  & \multicolumn{2}{c}{Accuracy$\uparrow$} & \multicolumn{2}{c}{BLEU$\uparrow$}          & \multicolumn{2}{c}{ROUGE-1$\uparrow$} & \multicolumn{2}{c}{ROUGE-2$\uparrow$} & \multicolumn{2}{c}{ROUGE-L$\uparrow$} \\
\midrule
LogDeduc  & \textcolor{Green}{0.500$\uparrow$} & 0.430 & - & - & - & - & - & - & - & - \\
QuestSel & \textcolor{Red}{0.545$\downarrow$} & 0.630 & - & - & - & - & - & - & - & - \\
StratQA & \textcolor{Green}{0.445$\uparrow$}         & 0.250        & 15.31 & \multicolumn{1}{l}{10.55} & \textcolor{Green}{0.195$\uparrow$}        & 0.135        & \textcolor{Green}{0.052$\uparrow$}        & 0.027        & \textcolor{Green}{0.182$\uparrow$}        & 0.128     \\
\bottomrule
\end{tabular}}
\end{table}

\begin{table}[ht]
\centering
\caption{The \textit{composite-5} evaluation results are presented in details with MeteoRA results on the left side and LoRA-B results on the right side of each metric column. A dash ('-') indicates that the corresponding metric was not applicable or included in the evaluation.}
\label{tab:serial_5_eval_results}
\resizebox{\columnwidth}{!}{
\begin{tabular}{ccccccccccc}
\toprule
Sub-task Name & \multicolumn{2}{c}{Accuracy$\uparrow$} & \multicolumn{2}{c}{BLEU$\uparrow$} & \multicolumn{2}{c}{ROUGE-1$\uparrow$} & \multicolumn{2}{c}{ROUGE-2$\uparrow$} & \multicolumn{2}{c}{ROUGE-L$\uparrow$} \\
\midrule
LogDeduc                 & 0.500 & 0.500 & -    & -    & -     & -     & -     & -     & -     & -     \\
QuestSel                & \textcolor{Red}{0.620$\downarrow$} & 0.770 & -    & -    & -     & -     & -     & -     & -     & -     \\
AbsNarr & \textcolor{Red}{0.350$\downarrow$} & 0.460 & -    & -    & -     & -     & -     & -     & -     & -     \\
GSWiki                & \textcolor{Green}{0.650$\uparrow$} & 0.410 & -    & -    & -     & -     & -     & -     & -     & -     \\
StratQA                         & \textcolor{Green}{0.495$\uparrow$} & 0.275 & \textcolor{Green}{9.86$\uparrow$} & 9.41 & \textcolor{Green}{0.221$\uparrow$} & 0.219 & \textcolor{Green}{0.069$\uparrow$} & 0.063 & \textcolor{Red}{0.207$\downarrow$} & 0.208 \\
\bottomrule
\end{tabular}}
\end{table}

\begin{table}[ht]
\centering
\caption{The \textit{composite-10} evaluation results are presented in details with MeteoRA results on the left side and LoRA-B results on the right side of each metric column. A dash ('-') indicates that the corresponding metric was not applicable or included in the evaluation. Note that the $0.00$ BLEU scores are caused by mismatch and too insufficient answers.}
\label{tab:serial_10_eval_results}
\resizebox{\columnwidth}{!}{
\begin{tabular}{ccccccccccc}
\toprule
Sub-task Name & \multicolumn{2}{c}{Accuracy$\uparrow$} & \multicolumn{2}{c}{BLEU$\uparrow$} & \multicolumn{2}{c}{ROUGE-1$\uparrow$} & \multicolumn{2}{c}{ROUGE-2$\uparrow$} & \multicolumn{2}{c}{ROUGE-L$\uparrow$} \\
\midrule
LogDeduc                 & \textcolor{Green}{0.500$\uparrow$} & 0.453 & -     & -     & -     & -     & -     & -     & -     & -     \\
QuestSel                & \textcolor{Green}{0.703$\uparrow$} & 0.688 & -     & -     & -     & -     & -     & -     & -     & -     \\
AbsNarr & \textcolor{Red}{0.625$\downarrow$} & 0.672 & -     & -     & -     & -     & -     & -     & -     & -     \\
GSWiki                & \textcolor{Green}{0.773$\uparrow$} & 0.727 & -     & -     & -     & -     & -     & -     & -     & -     \\
WinoWhy                            & \textcolor{Green}{0.422$\uparrow$} & 0.078 & -     & -     & -     & -     & -     & -     & -     & -     \\
StratQA                         & \textcolor{Green}{0.461$\uparrow$} & 0.211 & \textcolor{Green}{3.23$\uparrow$}  & 0.00  & \textcolor{Green}{0.225$\uparrow$} & 0.106 & \textcolor{Green}{0.051$\uparrow$} & 0.025 & \textcolor{Green}{0.210$\uparrow$} & 0.099 \\
DisflQA                          & \textcolor{Green}{0.266$\uparrow$} & 0.117 & -     & -     & -     & -     & -     & -     & -     & -     \\
NewsDE               & -     & -     & \textcolor{Green}{14.78$\uparrow$} & 14.54 & -     & -     & -     & -     & -     & -     \\
ALPACA                             & -     & -     & \textcolor{Red}{0.00$\downarrow$}  & 8.17  & \textcolor{Green}{0.257$\uparrow$} & 0.187 & 0.075 & 0.075 & \textcolor{Green}{0.241$\uparrow$} & 0.167 \\
LingPuzz               & -     & -     & \textcolor{Green}{17.37$\uparrow$} & 12.14 & \textcolor{Green}{0.233$\uparrow$} & 0.189 & \textcolor{Green}{0.052$\uparrow$} & 0.030 & \textcolor{Green}{0.176$\uparrow$} & 0.103 \\
\bottomrule
\end{tabular}}
\end{table}

\newpage
\subsection{Full content of the illustarted \textit{composite-n} task}\label{appendix:serial_3}

Below is the full content of a \textit{composite-3} task, which includes 2-shot examples, the input (3 problems), and the model generated output, shown in Figure \ref{fig:serial_3_short}.

\setlength{\fboxsep}{0pt}
\colorbox{NavyBlue!8}{\strut Example}\colorbox{NavyBlue!5}{\strut \ }\colorbox{gray!6}{\strut 1}\colorbox{gray!10}{\strut :$\backslash$n$\backslash$n}\colorbox{gray!21}{\strut [}\colorbox{gray!10}{\strut INST}\colorbox{gray!14}{\strut ]}\colorbox{Apricot!11}{\strut \ The}\colorbox{Apricot!10}{\strut \ following}\colorbox{Apricot!12}{\strut \ paragraphs}\colorbox{Apricot!19}{\strut \ each}\colorbox{Apricot!22}{\strut \ describe}\colorbox{Apricot!27}{\strut \ a}\colorbox{Apricot!12}{\strut \ set}\colorbox{Apricot!23}{\strut \ of}\colorbox{Apricot!31}{\strut \ five}\colorbox{Apricot!27}{\strut \ objects}\colorbox{Apricot!27}{\strut \ arranged}\colorbox{Apricot!29}{\strut \ in}\colorbox{Apricot!26}{\strut \ a}
\colorbox{Apricot!25}{\strut \ fixed}\colorbox{Apricot!30}{\strut \ order}\colorbox{Apricot!30}{\strut .}\colorbox{Apricot!36}{\strut \ The}\colorbox{Apricot!33}{\strut \ statements}\colorbox{Apricot!33}{\strut \ are}\colorbox{Apricot!30}{\strut \ logically}\colorbox{Apricot!31}{\strut \ consistent}\colorbox{Apricot!25}{\strut \ within}\colorbox{Apricot!30}{\strut \ each}\colorbox{Apricot!33}{\strut \ paragraph}\colorbox{Apricot!34}{\strut .$\backslash$n$\backslash$n}\colorbox{Apricot!31}{\strut On}\colorbox{Apricot!30}{\strut \ a}\colorbox{Apricot!22}{\strut \ branch}\colorbox{Apricot!31}{\strut ,}\colorbox{Apricot!37}{\strut \ there}
\colorbox{Apricot!39}{\strut \ are}\colorbox{Apricot!38}{\strut \ five}\colorbox{Apricot!36}{\strut \ birds}\colorbox{Apricot!41}{\strut :}\colorbox{Apricot!41}{\strut \ a}\colorbox{Apricot!37}{\strut \ cardinal}\colorbox{Apricot!41}{\strut ,}\colorbox{Apricot!39}{\strut \ a}\colorbox{Apricot!37}{\strut \ crow}\colorbox{Apricot!42}{\strut ,}\colorbox{Apricot!39}{\strut \ a}\colorbox{Apricot!35}{\strut \ fal}\colorbox{Apricot!37}{\strut con}\colorbox{Apricot!41}{\strut ,}\colorbox{Apricot!39}{\strut \ a}\colorbox{Apricot!39}{\strut \ robin}\colorbox{Apricot!41}{\strut ,}\colorbox{Apricot!39}{\strut \ and}\colorbox{Apricot!38}{\strut \ a}\colorbox{Apricot!34}{\strut \ blue}\colorbox{Apricot!33}{\strut \ j}\colorbox{Apricot!38}{\strut ay}\colorbox{Apricot!42}{\strut .}\colorbox{Apricot!43}{\strut \ The}\colorbox{Apricot!43}{\strut \ robin}\colorbox{Apricot!43}{\strut \ is}\colorbox{Apricot!39}{\strut \ to}\colorbox{Apricot!39}{\strut \ the}\colorbox{Apricot!39}{\strut \ right}\colorbox{Apricot!41}{\strut \ of}\colorbox{Apricot!40}{\strut \ the}
\colorbox{Apricot!40}{\strut \ cardinal}\colorbox{Apricot!43}{\strut .}\colorbox{Apricot!43}{\strut \ The}\colorbox{Apricot!42}{\strut \ cardinal}\colorbox{Apricot!43}{\strut \ is}\colorbox{Apricot!40}{\strut \ to}\colorbox{Apricot!36}{\strut \ the}\colorbox{Apricot!39}{\strut \ right}\colorbox{Apricot!41}{\strut \ of}\colorbox{Apricot!41}{\strut \ the}\colorbox{Apricot!39}{\strut \ blue}\colorbox{Apricot!32}{\strut \ j}\colorbox{Apricot!42}{\strut ay}\colorbox{Apricot!43}{\strut .}\colorbox{Apricot!43}{\strut \ The}\colorbox{Apricot!40}{\strut \ blue}\colorbox{Apricot!42}{\strut \ j}\colorbox{Apricot!42}{\strut ay}\colorbox{Apricot!43}{\strut \ is}\colorbox{Apricot!43}{\strut \ the}\colorbox{Apricot!43}{\strut \ second}\colorbox{Apricot!41}{\strut \ from}\colorbox{Apricot!39}{\strut \ the}\colorbox{Apricot!41}{\strut \ left}\colorbox{Apricot!43}{\strut .}\colorbox{Apricot!43}{\strut \ The}
\colorbox{Apricot!43}{\strut \ crow}\colorbox{Apricot!44}{\strut \ is}\colorbox{Apricot!43}{\strut \ the}\colorbox{Apricot!41}{\strut \ right}\colorbox{Apricot!42}{\strut most}\colorbox{Apricot!43}{\strut .}\colorbox{Apricot!39}{\strut \ Which}\colorbox{Apricot!36}{\strut \ choice}\colorbox{Apricot!39}{\strut \ is}\colorbox{Apricot!36}{\strut \ correct}\colorbox{Apricot!36}{\strut ?$\backslash$n}\colorbox{Apricot!33}{\strut choice}\colorbox{Apricot!38}{\strut :}\colorbox{Apricot!41}{\strut \ The}\colorbox{Apricot!42}{\strut \ cardinal}\colorbox{Apricot!43}{\strut \ is}\colorbox{Apricot!42}{\strut \ the}\colorbox{Apricot!41}{\strut \ left}\colorbox{Apricot!42}{\strut most}\colorbox{Apricot!39}{\strut .$\backslash$n}\colorbox{Apricot!37}{\strut choice}\colorbox{Apricot!40}{\strut :}
\colorbox{Apricot!41}{\strut \ The}\colorbox{Apricot!42}{\strut \ crow}\colorbox{Apricot!43}{\strut \ is}\colorbox{Apricot!42}{\strut \ the}\colorbox{Apricot!41}{\strut \ left}\colorbox{Apricot!42}{\strut most}\colorbox{Apricot!39}{\strut .$\backslash$n}\colorbox{Apricot!39}{\strut choice}\colorbox{Apricot!40}{\strut :}\colorbox{Apricot!41}{\strut \ The}\colorbox{Apricot!40}{\strut \ fal}\colorbox{Apricot!43}{\strut con}\colorbox{Apricot!43}{\strut \ is}\colorbox{Apricot!43}{\strut \ the}\colorbox{Apricot!42}{\strut \ left}\colorbox{Apricot!42}{\strut most}\colorbox{Apricot!40}{\strut .$\backslash$n}\colorbox{Apricot!35}{\strut choice}\colorbox{Apricot!40}{\strut :}\colorbox{Apricot!41}{\strut \ The}\colorbox{Apricot!43}{\strut \ robin}\colorbox{Apricot!43}{\strut \ is}\colorbox{Apricot!43}{\strut \ the}\colorbox{Apricot!42}{\strut \ left}\colorbox{Apricot!43}{\strut most}
\colorbox{Apricot!40}{\strut .$\backslash$n}\colorbox{Apricot!40}{\strut choice}\colorbox{Apricot!39}{\strut :}\colorbox{Apricot!41}{\strut \ The}\colorbox{Apricot!41}{\strut \ blue}\colorbox{Apricot!35}{\strut \ j}\colorbox{Apricot!42}{\strut ay}\colorbox{Apricot!43}{\strut \ is}\colorbox{Apricot!42}{\strut \ the}\colorbox{Apricot!42}{\strut \ left}\colorbox{Apricot!43}{\strut most}\colorbox{Apricot!40}{\strut .$\backslash$n}\colorbox{Apricot!12}{\strut [/}\colorbox{Apricot!29}{\strut INST}\colorbox{Apricot!33}{\strut ]$\backslash$n$\backslash$n}\colorbox{Apricot!15}{\strut [}\colorbox{Apricot!15}{\strut INST}\colorbox{Apricot!28}{\strut ]}\colorbox{Apricot!20}{\strut \ Given}\colorbox{Apricot!24}{\strut \ a}\colorbox{Aquamarine!25}{\strut \ short}\colorbox{Aquamarine!24}{\strut \ answer}\colorbox{Aquamarine!27}{\strut \ along}\colorbox{Aquamarine!24}{\strut \ with}\colorbox{Aquamarine!23}{\strut \ its}
\colorbox{Aquamarine!26}{\strut \ context}\colorbox{Aquamarine!26}{\strut ,}\colorbox{Aquamarine!22}{\strut \ select}\colorbox{Aquamarine!20}{\strut \ the}\colorbox{Aquamarine!19}{\strut \ most}\colorbox{Aquamarine!23}{\strut \ appropriate}\colorbox{Aquamarine!24}{\strut \ question}\colorbox{Aquamarine!24}{\strut \ which}\colorbox{Aquamarine!26}{\strut \ has}\colorbox{Aquamarine!25}{\strut \ the}\colorbox{Aquamarine!30}{\strut \ given}\colorbox{Aquamarine!29}{\strut \ short}\colorbox{Aquamarine!29}{\strut \ answer}\colorbox{Aquamarine!28}{\strut \ as}\colorbox{Aquamarine!25}{\strut \ its}\colorbox{Aquamarine!27}{\strut \ answer}\colorbox{Aquamarine!25}{\strut .$\backslash$n$\backslash$n}
\colorbox{Aquamarine!24}{\strut Here}\colorbox{Aquamarine!21}{\strut \ is}\colorbox{Aquamarine!25}{\strut \ the}\colorbox{Aquamarine!32}{\strut \ short}\colorbox{Aquamarine!31}{\strut \ answer}\colorbox{Aquamarine!26}{\strut \ followed}\colorbox{Aquamarine!26}{\strut \ by}\colorbox{Aquamarine!28}{\strut \ the}\colorbox{Aquamarine!26}{\strut \ context}\colorbox{Aquamarine!22}{\strut .$\backslash$n}\colorbox{Aquamarine!27}{\strut Short}\colorbox{Aquamarine!28}{\strut \ Answer}\colorbox{Apricot!22}{\strut :}\colorbox{Aquamarine!10}{\strut \ magazines}\colorbox{Aquamarine!14}{\strut \ and}\colorbox{Aquamarine!12}{\strut \ journals}\colorbox{Aquamarine!28}{\strut \ Context}\colorbox{Aquamarine!24}{\strut :}
\colorbox{Aquamarine!30}{\strut \ Tesla}\colorbox{Aquamarine!34}{\strut \ wrote}\colorbox{Aquamarine!32}{\strut \ a}\colorbox{Aquamarine!30}{\strut \ number}\colorbox{Aquamarine!31}{\strut \ of}\colorbox{Aquamarine!30}{\strut \ books}\colorbox{Aquamarine!30}{\strut \ and}\colorbox{Aquamarine!33}{\strut \ articles}\colorbox{Aquamarine!28}{\strut \ for}\colorbox{Aquamarine!23}{\strut \ magazines}\colorbox{Aquamarine!25}{\strut \ and}\colorbox{Aquamarine!28}{\strut \ journals}\colorbox{Aquamarine!37}{\strut .}\colorbox{Aquamarine!34}{\strut \ Among}\colorbox{Aquamarine!36}{\strut \ his}\colorbox{Aquamarine!33}{\strut \ books}\colorbox{Aquamarine!37}{\strut \ are}\colorbox{Aquamarine!29}{\strut \ My}
\colorbox{Aquamarine!24}{\strut \ In}\colorbox{Aquamarine!29}{\strut ventions}\colorbox{Aquamarine!32}{\strut :}\colorbox{Aquamarine!25}{\strut \ The}\colorbox{Aquamarine!27}{\strut \ Aut}\colorbox{Aquamarine!21}{\strut obi}\colorbox{Aquamarine!21}{\strut ography}\colorbox{Aquamarine!24}{\strut \ of}\colorbox{Aquamarine!26}{\strut \ Nikola}\colorbox{Aquamarine!28}{\strut \ Tesla}\colorbox{Aquamarine!37}{\strut ,}\colorbox{Aquamarine!30}{\strut \ compiled}\colorbox{Aquamarine!22}{\strut \ and}\colorbox{Aquamarine!28}{\strut \ edited}\colorbox{Aquamarine!28}{\strut \ by}\colorbox{Aquamarine!24}{\strut \ Ben}\colorbox{Aquamarine!27}{\strut \ Johnston}\colorbox{Aquamarine!37}{\strut ;}\colorbox{Aquamarine!32}{\strut \ The}
\colorbox{Aquamarine!22}{\strut \ Fantastic}\colorbox{Aquamarine!21}{\strut \ In}\colorbox{Aquamarine!19}{\strut ventions}\colorbox{Aquamarine!20}{\strut \ of}\colorbox{Aquamarine!28}{\strut \ Nikola}\colorbox{Aquamarine!25}{\strut \ Tesla}\colorbox{Aquamarine!30}{\strut ,}\colorbox{Aquamarine!28}{\strut \ compiled}\colorbox{Aquamarine!21}{\strut \ and}\colorbox{Aquamarine!26}{\strut \ edited}\colorbox{Aquamarine!23}{\strut \ by}\colorbox{Aquamarine!22}{\strut \ David}\colorbox{Aquamarine!16}{\strut \ H}\colorbox{Aquamarine!14}{\strut atcher}\colorbox{Aquamarine!20}{\strut \ Child}\colorbox{Aquamarine!29}{\strut ress}\colorbox{Aquamarine!36}{\strut ;}\colorbox{Aquamarine!33}{\strut \ and}\colorbox{Aquamarine!31}{\strut \ The}
\colorbox{Aquamarine!22}{\strut \ Tesla}\colorbox{Aquamarine!26}{\strut \ Papers}\colorbox{Aquamarine!37}{\strut .$\backslash$n}\colorbox{Aquamarine!25}{\strut choice}\colorbox{Aquamarine!35}{\strut :}\colorbox{Aquamarine!37}{\strut \ Who}\colorbox{Aquamarine!34}{\strut \ was}\colorbox{Aquamarine!32}{\strut \ the}\colorbox{Aquamarine!31}{\strut \ first}\colorbox{Aquamarine!30}{\strut \ to}\colorbox{Aquamarine!28}{\strut \ post}\colorbox{Aquamarine!32}{\strut \ tes}\colorbox{Aquamarine!32}{\strut la}\colorbox{Aquamarine!38}{\strut 's}\colorbox{Aquamarine!35}{\strut \ writings}\colorbox{Aquamarine!36}{\strut ?$\backslash$n}\colorbox{Aquamarine!29}{\strut choice}\colorbox{Aquamarine!39}{\strut :}\colorbox{Aquamarine!39}{\strut \ Who}\colorbox{Aquamarine!37}{\strut \ was}\colorbox{Aquamarine!30}{\strut \ in}\colorbox{Aquamarine!35}{\strut \ charge}\colorbox{Aquamarine!31}{\strut \ of}
\colorbox{Aquamarine!35}{\strut \ editing}\colorbox{Aquamarine!30}{\strut \ tes}\colorbox{Aquamarine!32}{\strut la}\colorbox{Aquamarine!34}{\strut 's}\colorbox{Aquamarine!36}{\strut \ autobiography}\colorbox{Aquamarine!38}{\strut ?$\backslash$n}\colorbox{Aquamarine!30}{\strut Choose}\colorbox{Aquamarine!32}{\strut \ the}\colorbox{Aquamarine!31}{\strut \ appropriate}\colorbox{Aquamarine!31}{\strut \ question}\colorbox{Aquamarine!32}{\strut \ which}\colorbox{Aquamarine!33}{\strut \ has}\colorbox{Aquamarine!37}{\strut \ the}\colorbox{Aquamarine!32}{\strut \ given}\colorbox{Aquamarine!34}{\strut \ answer}\colorbox{Aquamarine!32}{\strut .}\colorbox{Aquamarine!18}{\strut \ [/}
\colorbox{Aquamarine!18}{\strut INST}\colorbox{Aquamarine!29}{\strut ]$\backslash$n$\backslash$n}\colorbox{Aquamarine!10}{\strut [}\colorbox{Aquamarine!12}{\strut INST}\colorbox{Aquamarine!27}{\strut ]}\colorbox{Aquamarine!21}{\strut \ Q}\colorbox{Aquamarine!32}{\strut :}\colorbox{Aquamarine!26}{\strut \ Can}\colorbox{Magenta!28}{\strut \ Hulk}\colorbox{Magenta!29}{\strut 's}\colorbox{Magenta!14}{\strut \ alter}\colorbox{Magenta!29}{\strut \ ego}\colorbox{Magenta!21}{\strut \ explain}\colorbox{Magenta!17}{\strut \ atomic}\colorbox{Magenta!16}{\strut \ events}\colorbox{Aquamarine!26}{\strut ?$\backslash$n}\colorbox{Aquamarine!18}{\strut A}\colorbox{Aquamarine!18}{\strut :}\colorbox{Aquamarine!8}{\strut \ [/}\colorbox{Aquamarine!8}{\strut INST}\colorbox{Aquamarine!24}{\strut ]$\backslash$n$\backslash$n}\colorbox{Aquamarine!15}{\strut 1}\colorbox{Aquamarine!22}{\strut .}\colorbox{Aquamarine!18}{\strut \ The}\colorbox{Apricot!27}{\strut \ fal}\colorbox{Apricot!33}{\strut con}
\colorbox{Apricot!33}{\strut \ is}\colorbox{Apricot!35}{\strut \ the}\colorbox{Apricot!32}{\strut \ left}\colorbox{Apricot!34}{\strut most}\colorbox{Apricot!21}{\strut .$\backslash$n}\colorbox{Apricot!13}{\strut 2}\colorbox{Apricot!22}{\strut .}\colorbox{Aquamarine!29}{\strut \ Who}\colorbox{Aquamarine!29}{\strut \ was}\colorbox{Aquamarine!26}{\strut \ the}\colorbox{Aquamarine!29}{\strut \ first}\colorbox{Aquamarine!25}{\strut \ to}\colorbox{Aquamarine!32}{\strut \ post}\colorbox{Aquamarine!21}{\strut \ tes}\colorbox{Aquamarine!29}{\strut la}\colorbox{Aquamarine!28}{\strut 's}\colorbox{Aquamarine!33}{\strut \ writings}\colorbox{Aquamarine!25}{\strut ?$\backslash$n}\colorbox{Aquamarine!16}{\strut 3}\colorbox{Aquamarine!28}{\strut .}\colorbox{Aquamarine!16}{\strut \ Yes}\colorbox{Aquamarine!20}{\strut .}\colorbox{Aquamarine!17}{\strut \ Hulk}\colorbox{Aquamarine!22}{\strut 's}\colorbox{Aquamarine!24}{\strut \ alter}\colorbox{Aquamarine!21}{\strut \ ego}\colorbox{Magenta!18}{\strut \ is}\colorbox{Magenta!18}{\strut \ Dr}\colorbox{Magenta!16}{\strut .}
\colorbox{gray!11}{\strut \ Robert}\colorbox{Aquamarine!14}{\strut \ Bruce}\colorbox{Magenta!16}{\strut \ Banner}\colorbox{Magenta!21}{\strut .}\colorbox{Aquamarine!24}{\strut \ Dr}\colorbox{Aquamarine!18}{\strut .}\colorbox{Aquamarine!21}{\strut \ Robert}\colorbox{Aquamarine!17}{\strut \ Bruce}\colorbox{Magenta!18}{\strut \ Banner}\colorbox{Magenta!22}{\strut \ is}\colorbox{Magenta!23}{\strut \ a}\colorbox{Magenta!17}{\strut \ nuclear}\colorbox{Aquamarine!18}{\strut \ physicist}\colorbox{Magenta!21}{\strut .}\colorbox{Aquamarine!28}{\strut \ }\colorbox{Aquamarine!19}{\strut \ Nuclear}\colorbox{Magenta!20}{\strut \ physics}\colorbox{Magenta!21}{\strut \ is}\colorbox{Magenta!18}{\strut \ the}\colorbox{Magenta!16}{\strut \ field}
\colorbox{Magenta!15}{\strut \ of}\colorbox{Aquamarine!10}{\strut \ physics}\colorbox{Magenta!10}{\strut \ that}\colorbox{Magenta!9}{\strut \ studies}\colorbox{Magenta!13}{\strut \ atomic}\colorbox{gray!9}{\strut \ nuclei}\colorbox{Aquamarine!11}{\strut \ and}\colorbox{Aquamarine!9}{\strut \ their}\colorbox{gray!10}{\strut \ constituents}\colorbox{Aquamarine!10}{\strut \ and}\colorbox{Aquamarine!8}{\strut \ interactions}\colorbox{Aquamarine!23}{\strut .$\backslash$n$\backslash$n$\backslash$n}\colorbox{Aquamarine!9}{\strut Example}\colorbox{Aquamarine!17}{\strut \ }\colorbox{Aquamarine!6}{\strut 2}
\colorbox{Aquamarine!15}{\strut :$\backslash$n$\backslash$n}\colorbox{Aquamarine!8}{\strut [}\colorbox{Aquamarine!9}{\strut INST}\colorbox{Aquamarine!15}{\strut ]}\colorbox{Apricot!28}{\strut \ The}\colorbox{Apricot!23}{\strut \ following}\colorbox{Apricot!22}{\strut \ paragraphs}\colorbox{Apricot!16}{\strut \ each}\colorbox{Apricot!19}{\strut \ describe}\colorbox{Apricot!20}{\strut \ a}\colorbox{Aquamarine!11}{\strut \ set}\colorbox{Apricot!22}{\strut \ of}\colorbox{Apricot!31}{\strut \ three}\colorbox{Apricot!28}{\strut \ objects}\colorbox{Apricot!24}{\strut \ arranged}\colorbox{Apricot!22}{\strut \ in}\colorbox{Apricot!21}{\strut \ a}\colorbox{Apricot!19}{\strut \ fixed}\colorbox{Apricot!33}{\strut \ order}\colorbox{Apricot!26}{\strut .}
\colorbox{Apricot!24}{\strut \ The}\colorbox{Apricot!21}{\strut \ statements}\colorbox{Apricot!14}{\strut \ are}\colorbox{Apricot!15}{\strut \ logically}\colorbox{Apricot!16}{\strut \ consistent}\colorbox{Apricot!15}{\strut \ within}\colorbox{Apricot!23}{\strut \ each}\colorbox{Apricot!27}{\strut \ paragraph}\colorbox{Apricot!25}{\strut .$\backslash$n$\backslash$n}\colorbox{Apricot!26}{\strut In}\colorbox{Apricot!29}{\strut \ a}\colorbox{Apricot!22}{\strut \ golf}\colorbox{Apricot!24}{\strut \ tournament}\colorbox{Apricot!34}{\strut ,}\colorbox{Apricot!34}{\strut \ there}\colorbox{Apricot!29}{\strut \ were}
\colorbox{Apricot!31}{\strut \ three}\colorbox{Apricot!19}{\strut \ golf}\colorbox{Apricot!28}{\strut ers}\colorbox{Apricot!32}{\strut :}\colorbox{Apricot!22}{\strut \ Ana}\colorbox{Apricot!28}{\strut ,}\colorbox{Apricot!27}{\strut \ Rob}\colorbox{Apricot!34}{\strut ,}\colorbox{Apricot!29}{\strut \ and}\colorbox{Apricot!31}{\strut \ Joe}\colorbox{Apricot!36}{\strut .}\colorbox{Apricot!37}{\strut \ Joe}\colorbox{Apricot!35}{\strut \ finished}\colorbox{Apricot!35}{\strut \ above}\colorbox{Apricot!37}{\strut \ Ana}\colorbox{Apricot!37}{\strut .}\colorbox{Apricot!38}{\strut \ Rob}\colorbox{Apricot!40}{\strut \ finished}\colorbox{Apricot!35}{\strut \ above}\colorbox{Apricot!35}{\strut \ Joe}\colorbox{Apricot!37}{\strut .}\colorbox{Apricot!31}{\strut \ Which}\colorbox{Apricot!30}{\strut \ choice}\colorbox{Apricot!30}{\strut \ is}
\colorbox{Apricot!27}{\strut \ correct}\colorbox{Apricot!21}{\strut ?$\backslash$n}\colorbox{Apricot!31}{\strut choice}\colorbox{Apricot!37}{\strut :}\colorbox{Apricot!40}{\strut \ Ana}\colorbox{Apricot!40}{\strut \ finished}\colorbox{Apricot!40}{\strut \ second}\colorbox{Apricot!26}{\strut .$\backslash$n}\colorbox{Apricot!34}{\strut choice}\colorbox{Apricot!36}{\strut :}\colorbox{Apricot!39}{\strut \ Rob}\colorbox{Apricot!41}{\strut \ finished}\colorbox{Apricot!40}{\strut \ second}\colorbox{Apricot!27}{\strut .$\backslash$n}\colorbox{Apricot!31}{\strut choice}\colorbox{Apricot!36}{\strut :}\colorbox{Apricot!38}{\strut \ Joe}\colorbox{Apricot!41}{\strut \ finished}
\colorbox{Apricot!39}{\strut \ second}\colorbox{Apricot!31}{\strut .$\backslash$n}\colorbox{Apricot!11}{\strut [/}\colorbox{Apricot!15}{\strut INST}\colorbox{Apricot!18}{\strut ]$\backslash$n$\backslash$n}\colorbox{Apricot!10}{\strut [}\colorbox{Aquamarine!11}{\strut INST}\colorbox{Aquamarine!16}{\strut ]}\colorbox{Aquamarine!26}{\strut \ Given}\colorbox{Aquamarine!22}{\strut \ a}\colorbox{Aquamarine!24}{\strut \ short}\colorbox{Aquamarine!26}{\strut \ answer}\colorbox{Aquamarine!18}{\strut \ along}\colorbox{Aquamarine!23}{\strut \ with}\colorbox{Aquamarine!22}{\strut \ its}\colorbox{Aquamarine!20}{\strut \ context}\colorbox{Aquamarine!24}{\strut ,}\colorbox{Aquamarine!24}{\strut \ select}\colorbox{Aquamarine!21}{\strut \ the}\colorbox{Aquamarine!22}{\strut \ most}
\colorbox{Aquamarine!26}{\strut \ appropriate}\colorbox{Aquamarine!24}{\strut \ question}\colorbox{Aquamarine!19}{\strut \ which}\colorbox{Aquamarine!21}{\strut \ has}\colorbox{Aquamarine!16}{\strut \ the}\colorbox{Aquamarine!21}{\strut \ given}\colorbox{Aquamarine!19}{\strut \ short}\colorbox{Aquamarine!22}{\strut \ answer}\colorbox{Aquamarine!19}{\strut \ as}\colorbox{Apricot!21}{\strut \ its}\colorbox{Aquamarine!18}{\strut \ answer}\colorbox{Aquamarine!18}{\strut .$\backslash$n$\backslash$n}\colorbox{Aquamarine!14}{\strut Here}\colorbox{Aquamarine!15}{\strut \ is}\colorbox{Aquamarine!18}{\strut \ the}\colorbox{Aquamarine!23}{\strut \ short}\colorbox{Aquamarine!22}{\strut \ answer}
\colorbox{Aquamarine!16}{\strut \ followed}\colorbox{Aquamarine!14}{\strut \ by}\colorbox{Aquamarine!13}{\strut \ the}\colorbox{Aquamarine!17}{\strut \ context}\colorbox{Aquamarine!20}{\strut .$\backslash$n}\colorbox{Aquamarine!22}{\strut Short}\colorbox{Aquamarine!26}{\strut \ Answer}\colorbox{Aquamarine!22}{\strut :}\colorbox{Aquamarine!20}{\strut \ two}\colorbox{Aquamarine!16}{\strut \ months}\colorbox{Aquamarine!22}{\strut \ Context}\colorbox{Aquamarine!24}{\strut :}\colorbox{Aquamarine!21}{\strut \ It}\colorbox{Aquamarine!21}{\strut \ was}\colorbox{Aquamarine!22}{\strut \ not}\colorbox{Aquamarine!24}{\strut \ until}\colorbox{Aquamarine!21}{\strut \ January}\colorbox{Aquamarine!28}{\strut \ }\colorbox{Aquamarine!19}{\strut 151}\colorbox{Aquamarine!26}{\strut 8}\colorbox{Aquamarine!30}{\strut \ that}
\colorbox{Aquamarine!21}{\strut \ friends}\colorbox{Aquamarine!19}{\strut \ of}\colorbox{Aquamarine!29}{\strut \ Luther}\colorbox{Aquamarine!31}{\strut \ translated}\colorbox{Aquamarine!31}{\strut \ the}\colorbox{Aquamarine!31}{\strut \ }\colorbox{Aquamarine!26}{\strut 95}\colorbox{Aquamarine!23}{\strut \ Th}\colorbox{Aquamarine!26}{\strut eses}\colorbox{Aquamarine!24}{\strut \ from}\colorbox{Aquamarine!19}{\strut \ Latin}\colorbox{Aquamarine!28}{\strut \ into}\colorbox{Aquamarine!30}{\strut \ German}\colorbox{Aquamarine!31}{\strut \ and}\colorbox{Aquamarine!29}{\strut \ printed}\colorbox{Aquamarine!26}{\strut \ and}\colorbox{Aquamarine!23}{\strut \ widely}\colorbox{Aquamarine!28}{\strut \ copied}
\colorbox{Aquamarine!31}{\strut \ them}\colorbox{Aquamarine!37}{\strut ,}\colorbox{Aquamarine!31}{\strut \ making}\colorbox{Aquamarine!29}{\strut \ the}\colorbox{Aquamarine!29}{\strut \ controversy}\colorbox{Aquamarine!33}{\strut \ one}\colorbox{Aquamarine!30}{\strut \ of}\colorbox{Aquamarine!26}{\strut \ the}\colorbox{Aquamarine!23}{\strut \ first}\colorbox{Aquamarine!24}{\strut \ in}\colorbox{Aquamarine!26}{\strut \ history}\colorbox{Aquamarine!20}{\strut \ to}\colorbox{Aquamarine!22}{\strut \ be}\colorbox{Aquamarine!18}{\strut \ aided}\colorbox{Aquamarine!23}{\strut \ by}\colorbox{Aquamarine!23}{\strut \ the}\colorbox{Aquamarine!22}{\strut \ printing}\colorbox{Aquamarine!31}{\strut \ press}\colorbox{Aquamarine!38}{\strut .}\colorbox{Aquamarine!34}{\strut \ Within}
\colorbox{Aquamarine!29}{\strut \ two}\colorbox{Aquamarine!30}{\strut \ weeks}\colorbox{Aquamarine!34}{\strut ,}\colorbox{Aquamarine!23}{\strut \ copies}\colorbox{Aquamarine!29}{\strut \ of}\colorbox{Aquamarine!30}{\strut \ the}\colorbox{Aquamarine!31}{\strut \ the}\colorbox{Aquamarine!26}{\strut ses}\colorbox{Aquamarine!26}{\strut \ had}\colorbox{Aquamarine!20}{\strut \ spread}\colorbox{Aquamarine!23}{\strut \ throughout}\colorbox{Aquamarine!23}{\strut \ Germany}\colorbox{Aquamarine!36}{\strut ;}\colorbox{Aquamarine!27}{\strut \ within}\colorbox{Aquamarine!27}{\strut \ two}\colorbox{Aquamarine!24}{\strut \ months}\colorbox{Aquamarine!29}{\strut ,}\colorbox{Aquamarine!24}{\strut \ they}\colorbox{Aquamarine!19}{\strut \ had}
\colorbox{Aquamarine!21}{\strut \ spread}\colorbox{Aquamarine!24}{\strut \ throughout}\colorbox{Aquamarine!25}{\strut \ Europe}\colorbox{Aquamarine!30}{\strut .$\backslash$n}\colorbox{Aquamarine!26}{\strut choice}\colorbox{Aquamarine!35}{\strut :}\colorbox{Aquamarine!35}{\strut \ How}\colorbox{Aquamarine!29}{\strut \ long}\colorbox{Aquamarine!33}{\strut \ did}\colorbox{Aquamarine!31}{\strut \ the}\colorbox{Aquamarine!27}{\strut \ the}\colorbox{Aquamarine!24}{\strut ses}\colorbox{Aquamarine!26}{\strut \ take}\colorbox{Aquamarine!17}{\strut \ to}\colorbox{Aquamarine!25}{\strut \ spread}\colorbox{Aquamarine!22}{\strut \ through}\colorbox{Aquamarine!33}{\strut \ europe}\colorbox{Aquamarine!30}{\strut ?$\backslash$n}
\colorbox{Aquamarine!25}{\strut choice}\colorbox{Aquamarine!35}{\strut :}\colorbox{Aquamarine!34}{\strut \ How}\colorbox{Aquamarine!31}{\strut \ long}\colorbox{Aquamarine!33}{\strut \ did}\colorbox{Aquamarine!26}{\strut \ it}\colorbox{Aquamarine!27}{\strut \ take}\colorbox{Aquamarine!28}{\strut \ for}\colorbox{Aquamarine!32}{\strut \ the}\colorbox{Aquamarine!23}{\strut \ printing}\colorbox{Aquamarine!31}{\strut \ of}\colorbox{Aquamarine!33}{\strut \ the}\colorbox{Aquamarine!23}{\strut \ the}\colorbox{Aquamarine!29}{\strut ses}\colorbox{Aquamarine!27}{\strut \ to}\colorbox{Aquamarine!28}{\strut \ spread}\colorbox{Aquamarine!27}{\strut \ through}\colorbox{Aquamarine!36}{\strut \ germany}\colorbox{Aquamarine!30}{\strut ?$\backslash$n}\colorbox{Aquamarine!19}{\strut Choose}\colorbox{Aquamarine!25}{\strut \ the}
\colorbox{Aquamarine!28}{\strut \ appropriate}\colorbox{Aquamarine!24}{\strut \ question}\colorbox{Aquamarine!21}{\strut \ which}\colorbox{Aquamarine!25}{\strut \ has}\colorbox{Aquamarine!22}{\strut \ the}\colorbox{Aquamarine!23}{\strut \ given}\colorbox{Aquamarine!20}{\strut \ answer}\colorbox{Aquamarine!17}{\strut .}\colorbox{Aquamarine!10}{\strut \ [/}\colorbox{Aquamarine!11}{\strut INST}\colorbox{Aquamarine!17}{\strut ]$\backslash$n$\backslash$n}\colorbox{Aquamarine!9}{\strut [}\colorbox{Aquamarine!12}{\strut INST}\colorbox{Aquamarine!25}{\strut ]}\colorbox{Aquamarine!24}{\strut \ Q}\colorbox{Aquamarine!33}{\strut :}\colorbox{Aquamarine!24}{\strut \ Can}\colorbox{Magenta!21}{\strut \ Iowa}\colorbox{Magenta!20}{\strut \ be}\colorbox{Magenta!22}{\strut \ hidden}\colorbox{Magenta!22}{\strut \ in}
\colorbox{Magenta!18}{\strut \ the}\colorbox{Magenta!16}{\strut \ English}\colorbox{Magenta!25}{\strut \ Channel}\colorbox{Aquamarine!22}{\strut ?$\backslash$n}\colorbox{Aquamarine!15}{\strut A}\colorbox{Aquamarine!14}{\strut :}\colorbox{Aquamarine!10}{\strut \ [/}\colorbox{Aquamarine!11}{\strut INST}\colorbox{Aquamarine!14}{\strut ]$\backslash$n$\backslash$n}\colorbox{Aquamarine!13}{\strut 1}\colorbox{Aquamarine!20}{\strut .}\colorbox{Apricot!25}{\strut \ Joe}\colorbox{Apricot!31}{\strut \ finished}\colorbox{Apricot!33}{\strut \ second}\colorbox{Apricot!16}{\strut .$\backslash$n}\colorbox{Aquamarine!15}{\strut 2}\colorbox{Aquamarine!21}{\strut .}\colorbox{Aquamarine!26}{\strut \ How}\colorbox{Aquamarine!26}{\strut \ long}\colorbox{Aquamarine!27}{\strut \ did}\colorbox{Aquamarine!27}{\strut \ the}\colorbox{Aquamarine!15}{\strut \ the}\colorbox{Aquamarine!26}{\strut ses}\colorbox{Aquamarine!25}{\strut \ take}\colorbox{Aquamarine!20}{\strut \ to}
\colorbox{Aquamarine!22}{\strut \ spread}\colorbox{Aquamarine!24}{\strut \ through}\colorbox{Aquamarine!28}{\strut \ europe}\colorbox{Aquamarine!21}{\strut ?$\backslash$n}\colorbox{Aquamarine!14}{\strut 3}\colorbox{Aquamarine!23}{\strut .}\colorbox{Aquamarine!18}{\strut \ No}\colorbox{Magenta!23}{\strut .}\colorbox{Magenta!21}{\strut \ The}\colorbox{Magenta!23}{\strut \ maximum}\colorbox{Magenta!28}{\strut \ width}\colorbox{Magenta!30}{\strut \ of}\colorbox{Magenta!26}{\strut \ the}\colorbox{Aquamarine!19}{\strut \ English}\colorbox{Magenta!32}{\strut \ Channel}\colorbox{Magenta!32}{\strut \ is}\colorbox{Magenta!24}{\strut \ }\colorbox{Magenta!24}{\strut 150}\colorbox{Magenta!27}{\strut \ miles}\colorbox{Magenta!27}{\strut .}\colorbox{Magenta!28}{\strut \ The}
\colorbox{Magenta!27}{\strut \ minimum}\colorbox{Magenta!29}{\strut \ width}\colorbox{Magenta!30}{\strut \ of}\colorbox{Magenta!31}{\strut \ Iowa}\colorbox{Magenta!30}{\strut \ is}\colorbox{Magenta!21}{\strut \ }\colorbox{Magenta!23}{\strut 200}\colorbox{Magenta!20}{\strut \ miles}\colorbox{Aquamarine!14}{\strut .$\backslash$n$\backslash$n$\backslash$n}\colorbox{Aquamarine!17}{\strut Following}\colorbox{Aquamarine!22}{\strut \ the}\colorbox{Aquamarine!10}{\strut \ example}\colorbox{Plum!8}{\strut \ above}\colorbox{Aquamarine!15}{\strut ,}\colorbox{Aquamarine!11}{\strut \ generate}\colorbox{Aquamarine!10}{\strut \ answers}\colorbox{Aquamarine!11}{\strut \ to}\colorbox{Aquamarine!11}{\strut \ the}
\colorbox{Aquamarine!13}{\strut \ questions}\colorbox{Plum!8}{\strut \ below}\colorbox{Aquamarine!12}{\strut :$\backslash$n$\backslash$n}\colorbox{Aquamarine!5}{\strut [}\colorbox{Aquamarine!8}{\strut INST}\colorbox{Aquamarine!13}{\strut ]}\colorbox{Apricot!19}{\strut \ The}\colorbox{Apricot!18}{\strut \ following}\colorbox{Apricot!17}{\strut \ paragraphs}\colorbox{Apricot!14}{\strut \ each}\colorbox{Apricot!15}{\strut \ describe}\colorbox{Apricot!15}{\strut \ a}\colorbox{Cyan!8}{\strut \ set}\colorbox{Apricot!17}{\strut \ of}\colorbox{Apricot!27}{\strut \ seven}\colorbox{Apricot!24}{\strut \ objects}
\colorbox{Apricot!16}{\strut \ arranged}\colorbox{Apricot!15}{\strut \ in}\colorbox{Apricot!15}{\strut \ a}\colorbox{Apricot!15}{\strut \ fixed}\colorbox{Apricot!24}{\strut \ order}\colorbox{Apricot!17}{\strut .}\colorbox{Apricot!18}{\strut \ The}\colorbox{Apricot!18}{\strut \ statements}\colorbox{Apricot!11}{\strut \ are}\colorbox{Apricot!35}{\strut \ logically}\colorbox{Apricot!13}{\strut \ consistent}\colorbox{Apricot!13}{\strut \ within}\colorbox{Apricot!19}{\strut \ each}\colorbox{Apricot!19}{\strut \ paragraph}\colorbox{Apricot!22}{\strut .$\backslash$n$\backslash$n}\colorbox{Apricot!17}{\strut On}\colorbox{Apricot!21}{\strut \ a}
\colorbox{Apricot!22}{\strut \ branch}\colorbox{Apricot!22}{\strut ,}\colorbox{Apricot!19}{\strut \ there}\colorbox{Apricot!23}{\strut \ are}\colorbox{Apricot!29}{\strut \ seven}\colorbox{Apricot!29}{\strut \ birds}\colorbox{Apricot!35}{\strut :}\colorbox{Apricot!36}{\strut \ a}\colorbox{Apricot!36}{\strut \ cardinal}\colorbox{Apricot!36}{\strut ,}\colorbox{Apricot!36}{\strut \ a}\colorbox{Apricot!34}{\strut \ blue}\colorbox{Apricot!10}{\strut \ j}\colorbox{Apricot!37}{\strut ay}\colorbox{Apricot!38}{\strut ,}\colorbox{Apricot!37}{\strut \ a}\colorbox{Apricot!39}{\strut \ robin}\colorbox{Apricot!37}{\strut ,}\colorbox{Apricot!38}{\strut \ a}\colorbox{Apricot!30}{\strut \ fal}\colorbox{Apricot!39}{\strut con}\colorbox{Apricot!36}{\strut ,}\colorbox{Apricot!37}{\strut \ a}\colorbox{Apricot!30}{\strut \ qu}\colorbox{Apricot!34}{\strut ail}\colorbox{Apricot!35}{\strut ,}\colorbox{Apricot!36}{\strut \ a}\colorbox{Apricot!24}{\strut \ humming}\colorbox{Apricot!36}{\strut bird}\colorbox{Apricot!32}{\strut ,}\colorbox{Apricot!33}{\strut \ and}\colorbox{Apricot!34}{\strut \ a}
\colorbox{Apricot!30}{\strut \ r}\colorbox{Apricot!36}{\strut aven}\colorbox{Apricot!36}{\strut .}\colorbox{Apricot!38}{\strut \ The}\colorbox{Apricot!37}{\strut \ fal}\colorbox{Apricot!40}{\strut con}\colorbox{Apricot!42}{\strut \ is}\colorbox{Apricot!36}{\strut \ to}\colorbox{Apricot!37}{\strut \ the}\colorbox{Apricot!39}{\strut \ right}\colorbox{Apricot!39}{\strut \ of}\colorbox{Apricot!36}{\strut \ the}\colorbox{Apricot!36}{\strut \ qu}\colorbox{Apricot!39}{\strut ail}\colorbox{Apricot!39}{\strut .}\colorbox{Apricot!39}{\strut \ The}\colorbox{Apricot!42}{\strut \ robin}\colorbox{Apricot!42}{\strut \ is}\colorbox{Apricot!37}{\strut \ to}\colorbox{Apricot!23}{\strut \ the}\colorbox{Apricot!38}{\strut \ right}\colorbox{Apricot!39}{\strut \ of}\colorbox{Apricot!36}{\strut \ the}\colorbox{Apricot!33}{\strut \ humming}\colorbox{Apricot!39}{\strut bird}\colorbox{Apricot!38}{\strut .}\colorbox{Apricot!39}{\strut \ The}
\colorbox{Apricot!37}{\strut \ r}\colorbox{Apricot!40}{\strut aven}\colorbox{Apricot!42}{\strut \ is}\colorbox{Apricot!39}{\strut \ to}\colorbox{Apricot!27}{\strut \ the}\colorbox{Apricot!40}{\strut \ left}\colorbox{Apricot!40}{\strut \ of}\colorbox{Apricot!37}{\strut \ the}\colorbox{Apricot!38}{\strut \ qu}\colorbox{Apricot!38}{\strut ail}\colorbox{Apricot!39}{\strut .}\colorbox{Apricot!38}{\strut \ The}\colorbox{Apricot!38}{\strut \ blue}\colorbox{Apricot!34}{\strut \ j}\colorbox{Apricot!40}{\strut ay}\colorbox{Apricot!42}{\strut \ is}\colorbox{Apricot!41}{\strut \ the}\colorbox{Apricot!41}{\strut \ second}\colorbox{Apricot!39}{\strut \ from}\colorbox{Apricot!36}{\strut \ the}\colorbox{Apricot!40}{\strut \ left}\colorbox{Apricot!38}{\strut .}\colorbox{Apricot!38}{\strut \ The}\colorbox{Apricot!42}{\strut \ robin}\colorbox{Apricot!42}{\strut \ is}\colorbox{Apricot!40}{\strut \ the}\colorbox{Apricot!42}{\strut \ third}\colorbox{Apricot!38}{\strut \ from}
\colorbox{Apricot!38}{\strut \ the}\colorbox{Apricot!39}{\strut \ left}\colorbox{Apricot!38}{\strut .}\colorbox{Apricot!39}{\strut \ The}\colorbox{Apricot!41}{\strut \ cardinal}\colorbox{Apricot!42}{\strut \ is}\colorbox{Apricot!41}{\strut \ the}\colorbox{Apricot!41}{\strut \ third}\colorbox{Apricot!37}{\strut \ from}\colorbox{Apricot!23}{\strut \ the}\colorbox{Apricot!38}{\strut \ right}\colorbox{Apricot!36}{\strut .}\colorbox{Apricot!25}{\strut \ Which}\colorbox{Apricot!25}{\strut \ choice}\colorbox{Apricot!31}{\strut \ is}\colorbox{Apricot!22}{\strut \ correct}\colorbox{Apricot!15}{\strut ?$\backslash$n}\colorbox{Apricot!28}{\strut choice}\colorbox{Apricot!35}{\strut :}\colorbox{Apricot!37}{\strut \ The}\colorbox{Apricot!39}{\strut \ cardinal}\colorbox{Apricot!39}{\strut \ is}
\colorbox{Apricot!38}{\strut \ the}\colorbox{Apricot!40}{\strut \ second}\colorbox{Apricot!36}{\strut \ from}\colorbox{Apricot!37}{\strut \ the}\colorbox{Apricot!37}{\strut \ right}\colorbox{Apricot!21}{\strut .$\backslash$n}\colorbox{Apricot!28}{\strut choice}\colorbox{Apricot!36}{\strut :}\colorbox{Apricot!37}{\strut \ The}\colorbox{Apricot!37}{\strut \ blue}\colorbox{Aquamarine!11}{\strut \ j}\colorbox{Apricot!40}{\strut ay}\colorbox{Apricot!40}{\strut \ is}\colorbox{Apricot!39}{\strut \ the}\colorbox{Apricot!39}{\strut \ second}\colorbox{Apricot!37}{\strut \ from}\colorbox{Apricot!22}{\strut \ the}\colorbox{Apricot!33}{\strut \ right}\colorbox{Apricot!25}{\strut .$\backslash$n}\colorbox{Aquamarine!18}{\strut choice}\colorbox{Apricot!37}{\strut :}\colorbox{Apricot!36}{\strut \ The}\colorbox{Apricot!40}{\strut \ robin}
\colorbox{Apricot!39}{\strut \ is}\colorbox{Apricot!37}{\strut \ the}\colorbox{Apricot!39}{\strut \ second}\colorbox{Apricot!36}{\strut \ from}\colorbox{Apricot!18}{\strut \ the}\colorbox{Apricot!31}{\strut \ right}\colorbox{Apricot!22}{\strut .$\backslash$n}\colorbox{Apricot!12}{\strut choice}\colorbox{Apricot!36}{\strut :}\colorbox{Apricot!36}{\strut \ The}\colorbox{Apricot!38}{\strut \ fal}\colorbox{Apricot!41}{\strut con}\colorbox{Apricot!40}{\strut \ is}\colorbox{Apricot!39}{\strut \ the}\colorbox{Apricot!39}{\strut \ second}\colorbox{Apricot!36}{\strut \ from}\colorbox{Apricot!25}{\strut \ the}\colorbox{Apricot!30}{\strut \ right}\colorbox{Apricot!23}{\strut .$\backslash$n}\colorbox{Apricot!30}{\strut choice}\colorbox{Apricot!35}{\strut :}\colorbox{Apricot!36}{\strut \ The}\colorbox{Apricot!37}{\strut \ qu}\colorbox{Apricot!39}{\strut ail}
\colorbox{Apricot!40}{\strut \ is}\colorbox{Apricot!39}{\strut \ the}\colorbox{Apricot!39}{\strut \ second}\colorbox{Apricot!36}{\strut \ from}\colorbox{Apricot!32}{\strut \ the}\colorbox{Apricot!33}{\strut \ right}\colorbox{Apricot!23}{\strut .$\backslash$n}\colorbox{Aquamarine!21}{\strut choice}\colorbox{Apricot!35}{\strut :}\colorbox{Apricot!35}{\strut \ The}\colorbox{Apricot!34}{\strut \ humming}\colorbox{Apricot!39}{\strut bird}\colorbox{Apricot!39}{\strut \ is}\colorbox{Apricot!39}{\strut \ the}\colorbox{Apricot!39}{\strut \ second}\colorbox{Apricot!35}{\strut \ from}\colorbox{Aquamarine!17}{\strut \ the}\colorbox{Apricot!32}{\strut \ right}\colorbox{Apricot!24}{\strut .$\backslash$n}\colorbox{Apricot!31}{\strut choice}\colorbox{Apricot!35}{\strut :}\colorbox{Apricot!36}{\strut \ The}
\colorbox{Apricot!36}{\strut \ r}\colorbox{Apricot!39}{\strut aven}\colorbox{Apricot!39}{\strut \ is}\colorbox{Apricot!38}{\strut \ the}\colorbox{Apricot!39}{\strut \ second}\colorbox{Apricot!36}{\strut \ from}\colorbox{Apricot!17}{\strut \ the}\colorbox{Apricot!32}{\strut \ right}\colorbox{Apricot!21}{\strut .$\backslash$n}\colorbox{Apricot!11}{\strut \ [/}\colorbox{Apricot!7}{\strut INST}\colorbox{Apricot!8}{\strut ]$\backslash$n$\backslash$n$\backslash$n}\colorbox{gray!5}{\strut [}\colorbox{Apricot!7}{\strut INST}\colorbox{Aquamarine!11}{\strut ]}\colorbox{Aquamarine!22}{\strut \ Given}\colorbox{Aquamarine!20}{\strut \ a}\colorbox{Aquamarine!17}{\strut \ short}\colorbox{Aquamarine!21}{\strut \ answer}\colorbox{Aquamarine!14}{\strut \ along}\colorbox{Aquamarine!16}{\strut \ with}\colorbox{Aquamarine!17}{\strut \ its}
\colorbox{Aquamarine!17}{\strut \ context}\colorbox{Aquamarine!16}{\strut ,}\colorbox{Aquamarine!16}{\strut \ select}\colorbox{Aquamarine!15}{\strut \ the}\colorbox{Cyan!11}{\strut \ most}\colorbox{Aquamarine!21}{\strut \ appropriate}\colorbox{Aquamarine!16}{\strut \ question}\colorbox{Aquamarine!13}{\strut \ which}\colorbox{Aquamarine!16}{\strut \ has}\colorbox{Aquamarine!13}{\strut \ the}\colorbox{Aquamarine!16}{\strut \ given}\colorbox{Aquamarine!16}{\strut \ short}\colorbox{Aquamarine!17}{\strut \ answer}\colorbox{Aquamarine!13}{\strut \ as}\colorbox{Aquamarine!13}{\strut \ its}\colorbox{Aquamarine!10}{\strut \ answer}\colorbox{Aquamarine!12}{\strut .$\backslash$n$\backslash$n}
\colorbox{Aquamarine!10}{\strut Here}\colorbox{Aquamarine!12}{\strut \ is}\colorbox{Aquamarine!12}{\strut \ the}\colorbox{Aquamarine!18}{\strut \ short}\colorbox{Aquamarine!17}{\strut \ answer}\colorbox{Aquamarine!11}{\strut \ followed}\colorbox{Aquamarine!11}{\strut \ by}\colorbox{Apricot!12}{\strut \ the}\colorbox{Aquamarine!14}{\strut \ context}\colorbox{Aquamarine!12}{\strut .$\backslash$n}\colorbox{Aquamarine!16}{\strut Short}\colorbox{Aquamarine!18}{\strut \ Answer}\colorbox{Aquamarine!19}{\strut :}\colorbox{Aquamarine!17}{\strut \ Gh}\colorbox{Aquamarine!11}{\strut az}\colorbox{Aquamarine!18}{\strut an}\colorbox{Aquamarine!26}{\strut \ Khan}\colorbox{Aquamarine!20}{\strut \ Context}\colorbox{Aquamarine!30}{\strut :}\colorbox{Aquamarine!35}{\strut \ The}
\colorbox{Aquamarine!30}{\strut \ inv}\colorbox{Aquamarine!28}{\strut asions}\colorbox{Aquamarine!29}{\strut \ of}\colorbox{Aquamarine!25}{\strut \ Baghdad}\colorbox{Aquamarine!33}{\strut ,}\colorbox{Aquamarine!22}{\strut \ Sam}\colorbox{Aquamarine!12}{\strut ark}\colorbox{Aquamarine!24}{\strut and}\colorbox{Aquamarine!30}{\strut ,}\colorbox{Aquamarine!21}{\strut \ Urg}\colorbox{Aquamarine!25}{\strut ench}\colorbox{Aquamarine!32}{\strut ,}\colorbox{Aquamarine!30}{\strut \ Kiev}\colorbox{Aquamarine!34}{\strut ,}\colorbox{Aquamarine!26}{\strut \ Vladimir}\colorbox{Aquamarine!32}{\strut \ among}\colorbox{Aquamarine!32}{\strut \ others}\colorbox{Aquamarine!34}{\strut \ caused}\colorbox{Aquamarine!30}{\strut \ mass}\colorbox{Aquamarine!29}{\strut \ murders}\colorbox{Aquamarine!33}{\strut ,}
\colorbox{Aquamarine!36}{\strut \ such}\colorbox{Aquamarine!33}{\strut \ as}\colorbox{Aquamarine!33}{\strut \ when}\colorbox{Aquamarine!25}{\strut \ portions}\colorbox{Aquamarine!27}{\strut \ of}\colorbox{Aquamarine!31}{\strut \ southern}\colorbox{Aquamarine!26}{\strut \ Kh}\colorbox{Aquamarine!18}{\strut uz}\colorbox{Aquamarine!18}{\strut est}\colorbox{Aquamarine!27}{\strut an}\colorbox{Aquamarine!31}{\strut \ were}\colorbox{Aquamarine!24}{\strut \ completely}\colorbox{Aquamarine!31}{\strut \ destroyed}\colorbox{Aquamarine!37}{\strut .}\colorbox{Aquamarine!34}{\strut \ His}\colorbox{Aquamarine!31}{\strut \ descendant}\colorbox{Aquamarine!24}{\strut \ H}\colorbox{Aquamarine!20}{\strut ul}\colorbox{Aquamarine!26}{\strut agu}
\colorbox{Aquamarine!32}{\strut \ Khan}\colorbox{Aquamarine!33}{\strut \ destroyed}\colorbox{Aquamarine!34}{\strut \ much}\colorbox{Aquamarine!28}{\strut \ of}\colorbox{Aquamarine!29}{\strut \ Iran}\colorbox{Aquamarine!32}{\strut 's}\colorbox{Aquamarine!27}{\strut \ northern}\colorbox{Aquamarine!36}{\strut \ part}\colorbox{Aquamarine!36}{\strut \ and}\colorbox{Aquamarine!30}{\strut \ sacked}\colorbox{Aquamarine!27}{\strut \ Baghdad}\colorbox{Aquamarine!35}{\strut \ although}\colorbox{Aquamarine!33}{\strut \ his}\colorbox{Aquamarine!29}{\strut \ forces}\colorbox{Aquamarine!29}{\strut \ were}\colorbox{Aquamarine!30}{\strut \ halted}
\colorbox{Aquamarine!30}{\strut \ by}\colorbox{Aquamarine!31}{\strut \ the}\colorbox{Aquamarine!31}{\strut \ M}\colorbox{Aquamarine!24}{\strut aml}\colorbox{Aquamarine!23}{\strut u}\colorbox{Aquamarine!30}{\strut ks}\colorbox{Aquamarine!30}{\strut \ of}\colorbox{Aquamarine!33}{\strut \ Egypt}\colorbox{Aquamarine!39}{\strut ,}\colorbox{Aquamarine!35}{\strut \ but}\colorbox{Aquamarine!31}{\strut \ H}\colorbox{Aquamarine!26}{\strut ul}\colorbox{Aquamarine!30}{\strut agu}\colorbox{Aquamarine!35}{\strut 's}\colorbox{Aquamarine!32}{\strut \ descendant}\colorbox{Aquamarine!23}{\strut \ Gh}\colorbox{Aquamarine!15}{\strut az}\colorbox{Aquamarine!23}{\strut an}\colorbox{Aquamarine!27}{\strut \ Khan}\colorbox{Aquamarine!34}{\strut \ would}\colorbox{Aquamarine!31}{\strut \ return}\colorbox{Aquamarine!28}{\strut \ to}\colorbox{Aquamarine!27}{\strut \ beat}\colorbox{Aquamarine!27}{\strut \ the}\colorbox{Aquamarine!30}{\strut \ Egyptian}
\colorbox{Aquamarine!34}{\strut \ M}\colorbox{Aquamarine!26}{\strut aml}\colorbox{Aquamarine!18}{\strut u}\colorbox{Aquamarine!28}{\strut ks}\colorbox{Aquamarine!27}{\strut \ right}\colorbox{Aquamarine!23}{\strut \ out}\colorbox{Aquamarine!24}{\strut \ of}\colorbox{Aquamarine!23}{\strut \ Lev}\colorbox{Aquamarine!28}{\strut ant}\colorbox{Aquamarine!36}{\strut ,}\colorbox{Aquamarine!28}{\strut \ Palestine}\colorbox{Aquamarine!31}{\strut \ and}\colorbox{Aquamarine!30}{\strut \ even}\colorbox{Aquamarine!27}{\strut \ Gaza}\colorbox{Aquamarine!33}{\strut .}\colorbox{Aquamarine!34}{\strut \ According}\colorbox{Aquamarine!33}{\strut \ to}\colorbox{Aquamarine!32}{\strut \ the}\colorbox{Aquamarine!29}{\strut \ works}\colorbox{Aquamarine!26}{\strut \ of}\colorbox{Aquamarine!28}{\strut \ the}\colorbox{Aquamarine!25}{\strut \ Persian}
\colorbox{Aquamarine!22}{\strut \ historian}\colorbox{Aquamarine!18}{\strut \ Rash}\colorbox{gray!15}{\strut id}\colorbox{Aquamarine!12}{\strut -al}\colorbox{Aquamarine!14}{\strut -D}\colorbox{Aquamarine!15}{\strut in}\colorbox{Aquamarine!18}{\strut \ Ham}\colorbox{Aquamarine!12}{\strut ad}\colorbox{Aquamarine!26}{\strut ani}\colorbox{Aquamarine!35}{\strut ,}\colorbox{Aquamarine!35}{\strut \ the}\colorbox{Aquamarine!29}{\strut \ Mong}\colorbox{Aquamarine!33}{\strut ols}\colorbox{Aquamarine!27}{\strut \ killed}\colorbox{Aquamarine!27}{\strut \ more}\colorbox{Aquamarine!23}{\strut \ than}\colorbox{Aquamarine!20}{\strut \ }\colorbox{Aquamarine!30}{\strut 70}\colorbox{Aquamarine!24}{\strut ,}\colorbox{Aquamarine!27}{\strut 000}\colorbox{Aquamarine!23}{\strut \ people}\colorbox{Aquamarine!32}{\strut \ in}\colorbox{Aquamarine!30}{\strut \ M}\colorbox{Aquamarine!21}{\strut erv}\colorbox{Aquamarine!30}{\strut \ and}\colorbox{Aquamarine!33}{\strut \ more}
\colorbox{Aquamarine!24}{\strut \ than}\colorbox{Aquamarine!21}{\strut \ }\colorbox{Aquamarine!26}{\strut 190}\colorbox{Aquamarine!29}{\strut ,}\colorbox{Aquamarine!23}{\strut 000}\colorbox{Aquamarine!25}{\strut \ in}\colorbox{Aquamarine!23}{\strut \ N}\colorbox{Aquamarine!17}{\strut ish}\colorbox{Aquamarine!19}{\strut apur}\colorbox{Aquamarine!36}{\strut .}\colorbox{Aquamarine!37}{\strut \ In}\colorbox{Aquamarine!32}{\strut \ }\colorbox{Aquamarine!23}{\strut 123}\colorbox{Aquamarine!31}{\strut 7}\colorbox{Aquamarine!24}{\strut \ Bat}\colorbox{Aquamarine!23}{\strut u}\colorbox{Aquamarine!29}{\strut \ Khan}\colorbox{Aquamarine!34}{\strut ,}\colorbox{Aquamarine!29}{\strut \ a}\colorbox{Aquamarine!22}{\strut \ grandson}\colorbox{Aquamarine!24}{\strut \ of}\colorbox{Aquamarine!26}{\strut \ G}\colorbox{Aquamarine!27}{\strut eng}\colorbox{Aquamarine!25}{\strut his}\colorbox{Aquamarine!27}{\strut \ Khan}\colorbox{Aquamarine!32}{\strut ,}\colorbox{Aquamarine!29}{\strut \ launched}\colorbox{Aquamarine!28}{\strut \ an}\colorbox{Aquamarine!25}{\strut \ invasion}
\colorbox{Aquamarine!28}{\strut \ into}\colorbox{Aquamarine!27}{\strut \ K}\colorbox{Aquamarine!24}{\strut ie}\colorbox{Aquamarine!24}{\strut van}\colorbox{Aquamarine!25}{\strut \ Rus}\colorbox{Aquamarine!35}{\strut '.}\colorbox{Aquamarine!27}{\strut \ Over}\colorbox{Aquamarine!30}{\strut \ the}\colorbox{Aquamarine!28}{\strut \ course}\colorbox{Aquamarine!25}{\strut \ of}\colorbox{Aquamarine!29}{\strut \ three}\colorbox{Aquamarine!24}{\strut \ years}\colorbox{Aquamarine!28}{\strut ,}\colorbox{Aquamarine!29}{\strut \ the}\colorbox{Aquamarine!26}{\strut \ Mong}\colorbox{Aquamarine!23}{\strut ols}\colorbox{Aquamarine!26}{\strut \ destroyed}\colorbox{Aquamarine!27}{\strut \ and}\colorbox{Aquamarine!21}{\strut \ annihil}\colorbox{Aquamarine!23}{\strut ated}\colorbox{Aquamarine!26}{\strut \ all}\colorbox{Aquamarine!27}{\strut \ of}\colorbox{Aquamarine!26}{\strut \ the}
\colorbox{Aquamarine!26}{\strut \ major}\colorbox{Aquamarine!21}{\strut \ cities}\colorbox{Aquamarine!27}{\strut \ of}\colorbox{Aquamarine!25}{\strut \ Eastern}\colorbox{Aquamarine!28}{\strut \ Europe}\colorbox{Aquamarine!31}{\strut \ with}\colorbox{Aquamarine!28}{\strut \ the}\colorbox{Aquamarine!26}{\strut \ exceptions}\colorbox{Aquamarine!25}{\strut \ of}\colorbox{Aquamarine!22}{\strut \ Nov}\colorbox{Aquamarine!17}{\strut gor}\colorbox{Aquamarine!26}{\strut od}\colorbox{Aquamarine!30}{\strut \ and}\colorbox{Aquamarine!28}{\strut \ P}\colorbox{gray!17}{\strut sk}\colorbox{Aquamarine!21}{\strut ov}\colorbox{Aquamarine!29}{\strut .$\backslash$n}\colorbox{Aquamarine!22}{\strut \ }\colorbox{Aquamarine!15}{\strut \ choice}\colorbox{Aquamarine!31}{\strut :}\colorbox{Aquamarine!37}{\strut \ Which}
\colorbox{Aquamarine!33}{\strut \ g}\colorbox{Aquamarine!31}{\strut eng}\colorbox{Aquamarine!30}{\strut his}\colorbox{Aquamarine!32}{\strut \ k}\colorbox{Aquamarine!32}{\strut han}\colorbox{Aquamarine!35}{\strut \ descendant}\colorbox{Aquamarine!35}{\strut \ sacked}\colorbox{Aquamarine!25}{\strut \ bag}\colorbox{Aquamarine!23}{\strut hd}\colorbox{Aquamarine!35}{\strut ad}\colorbox{Aquamarine!23}{\strut ?$\backslash$n}\colorbox{Aquamarine!16}{\strut \ }\colorbox{Aquamarine!22}{\strut \ choice}\colorbox{Aquamarine!36}{\strut :}\colorbox{Aquamarine!38}{\strut \ Which}\colorbox{Aquamarine!34}{\strut \ of}\colorbox{Aquamarine!32}{\strut \ eastern}\colorbox{Aquamarine!37}{\strut \ europe}\colorbox{Aquamarine!39}{\strut 's}\colorbox{Aquamarine!29}{\strut \ big}\colorbox{Aquamarine!34}{\strut \ cities}\colorbox{Aquamarine!35}{\strut \ survived}
\colorbox{Aquamarine!30}{\strut \ the}\colorbox{Aquamarine!32}{\strut \ mong}\colorbox{Aquamarine!30}{\strut ol}\colorbox{Aquamarine!36}{\strut \ invasion}\colorbox{Aquamarine!24}{\strut ?$\backslash$n}\colorbox{Aquamarine!20}{\strut \ }\colorbox{Aquamarine!27}{\strut \ choice}\colorbox{Aquamarine!37}{\strut :}\colorbox{Aquamarine!39}{\strut \ Which}\colorbox{Aquamarine!36}{\strut \ of}\colorbox{Aquamarine!32}{\strut \ g}\colorbox{Aquamarine!19}{\strut eng}\colorbox{Aquamarine!31}{\strut his}\colorbox{Aquamarine!14}{\strut \ k}\colorbox{Aquamarine!31}{\strut han}\colorbox{Aquamarine!33}{\strut 's}\colorbox{Aquamarine!36}{\strut \ descendants}\colorbox{Aquamarine!36}{\strut \ was}\colorbox{Aquamarine!37}{\strut \ responsible}\colorbox{Aquamarine!36}{\strut \ for}\colorbox{Aquamarine!32}{\strut \ driving}
\colorbox{Aquamarine!34}{\strut \ the}\colorbox{Aquamarine!34}{\strut \ m}\colorbox{Aquamarine!28}{\strut aml}\colorbox{Aquamarine!25}{\strut u}\colorbox{Aquamarine!29}{\strut ks}\colorbox{Aquamarine!34}{\strut \ from}\colorbox{Aquamarine!26}{\strut \ pale}\colorbox{Aquamarine!24}{\strut st}\colorbox{Aquamarine!33}{\strut ine}\colorbox{Aquamarine!24}{\strut ?$\backslash$n}\colorbox{Aquamarine!13}{\strut Choose}\colorbox{Aquamarine!19}{\strut \ the}\colorbox{Aquamarine!20}{\strut \ appropriate}\colorbox{Aquamarine!19}{\strut \ question}\colorbox{Aquamarine!14}{\strut \ which}\colorbox{Aquamarine!20}{\strut \ has}\colorbox{Aquamarine!24}{\strut \ the}\colorbox{Aquamarine!19}{\strut \ given}\colorbox{Aquamarine!16}{\strut \ answer}\colorbox{Aquamarine!10}{\strut .}\colorbox{Aquamarine!8}{\strut \ [/}
\colorbox{gray!4}{\strut INST}\colorbox{Aquamarine!10}{\strut ]$\backslash$n$\backslash$n$\backslash$n}\colorbox{gray!6}{\strut [}\colorbox{Aquamarine!6}{\strut INST}\colorbox{Aquamarine!15}{\strut ]}\colorbox{Aquamarine!11}{\strut \ Q}\colorbox{Aquamarine!20}{\strut :}\colorbox{Aquamarine!19}{\strut \ Could}\colorbox{Aquamarine!22}{\strut \ the}\colorbox{Aquamarine!20}{\strut \ main}\colorbox{Magenta!22}{\strut \ character}\colorbox{Magenta!17}{\strut \ of}\colorbox{Magenta!11}{\strut \ "}\colorbox{Magenta!9}{\strut Alice}\colorbox{Aquamarine!8}{\strut 's}\colorbox{Magenta!12}{\strut \ Adventures}\colorbox{Aquamarine!10}{\strut \ in}\colorbox{Magenta!20}{\strut \ Wonderland}\colorbox{Magenta!27}{\strut "}\colorbox{Magenta!29}{\strut \ join}\colorbox{Magenta!26}{\strut \ a}
\colorbox{Magenta!24}{\strut \ Mason}\colorbox{Magenta!27}{\strut ic}\colorbox{Magenta!31}{\strut \ Lodge}\colorbox{Aquamarine!14}{\strut ?$\backslash$n}\colorbox{Aquamarine!9}{\strut A}\colorbox{Aquamarine!11}{\strut :}\colorbox{Aquamarine!6}{\strut \ [/}\colorbox{gray!6}{\strut INST}\colorbox{Aquamarine!8}{\strut ]$\backslash$n$\backslash$n}\colorbox{Apricot!11}{\strut 1}\colorbox{Aquamarine!11}{\strut .}\colorbox{Apricot!23}{\strut \ The}\colorbox{Apricot!26}{\strut \ qu}\colorbox{Apricot!31}{\strut ail}\colorbox{Apricot!30}{\strut \ is}\colorbox{Apricot!32}{\strut \ the}\colorbox{Apricot!33}{\strut \ second}\colorbox{Apricot!30}{\strut \ from}\colorbox{Magenta!14}{\strut \ the}\colorbox{Apricot!21}{\strut \ right}\colorbox{Apricot!10}{\strut .$\backslash$n}\colorbox{Aquamarine!8}{\strut 2}\colorbox{Aquamarine!17}{\strut .}\colorbox{Aquamarine!21}{\strut \ Which}\colorbox{Aquamarine!20}{\strut \ of}\colorbox{Aquamarine!20}{\strut \ g}\colorbox{Aquamarine!22}{\strut eng}\colorbox{Aquamarine!27}{\strut his}
\colorbox{Aquamarine!13}{\strut \ k}\colorbox{Aquamarine!28}{\strut han}\colorbox{Aquamarine!28}{\strut 's}\colorbox{Aquamarine!29}{\strut \ descendants}\colorbox{Aquamarine!28}{\strut \ was}\colorbox{Aquamarine!29}{\strut \ responsible}\colorbox{Aquamarine!29}{\strut \ for}\colorbox{Aquamarine!25}{\strut \ driving}\colorbox{Aquamarine!26}{\strut \ the}\colorbox{Aquamarine!13}{\strut \ m}\colorbox{Aquamarine!28}{\strut aml}\colorbox{Aquamarine!20}{\strut u}\colorbox{Aquamarine!29}{\strut ks}\colorbox{Aquamarine!30}{\strut \ from}\colorbox{Aquamarine!20}{\strut \ pale}\colorbox{Aquamarine!9}{\strut st}\colorbox{Aquamarine!25}{\strut ine}\colorbox{Aquamarine!13}{\strut ?$\backslash$n}\colorbox{Aquamarine!7}{\strut 3}\colorbox{Aquamarine!14}{\strut .}\colorbox{Magenta!14}{\strut \ No}\colorbox{Magenta!27}{\strut .}\colorbox{Magenta!26}{\strut \ The}\colorbox{Aquamarine!18}{\strut \ main}
\colorbox{Magenta!28}{\strut \ character}\colorbox{Magenta!21}{\strut \ of}\colorbox{Magenta!13}{\strut \ "}\colorbox{Magenta!12}{\strut Alice}\colorbox{Magenta!17}{\strut 's}\colorbox{Magenta!16}{\strut \ Adventures}\colorbox{Aquamarine!7}{\strut \ in}\colorbox{Magenta!22}{\strut \ Wonderland}\colorbox{Magenta!25}{\strut "}\colorbox{Magenta!29}{\strut \ is}\colorbox{Magenta!23}{\strut \ Alice}\colorbox{Magenta!30}{\strut .}\colorbox{Magenta!31}{\strut \ Women}\colorbox{Magenta!33}{\strut \ are}\colorbox{Magenta!32}{\strut \ not}\colorbox{Magenta!32}{\strut \ allowed}\colorbox{Magenta!33}{\strut \ to}\colorbox{Magenta!34}{\strut \ join}\colorbox{Magenta!30}{\strut \ Mason}\colorbox{Magenta!33}{\strut ic}
\colorbox{Magenta!24}{\strut \ Lod}\colorbox{Magenta!32}{\strut ges}\colorbox{Magenta!23}{\strut .}

\colorbox{Aquamarine!30}{\strut \hspace{1.1cm}} \textit{logical\_deduction}
\hspace{0.2cm}
\colorbox{Apricot!30}{\strut \hspace{1.1cm}} \textit{question\_selection}
\hspace{0.2cm}
\colorbox{Magenta!30}{\strut \hspace{1.1cm}} \textit{strategyqa}
\hspace{0.2cm}
\colorbox{gray!10}{\strut \hspace{1.1cm}} other task

\newpage
\subsection{Efficiency evaluation experiments on different MeteoRA forward pass implementations}\label{appendix: efcy_exps}

In addition to experiments on our 28 selected tasks, we assess the efficiency of our MeteoRA forward pass design using randomly-generated pseudo data across various settings, including batch size ($b$), sequence length ($s$), gating weights top-k ($k$), LoRA rank size ($r$), number of LoRAs ($l$), maximum tokens to generate ($g$), input hidden dimension ($h$), and output hidden dimension ($hout$). Moreover, here we introduce a new baseline, \textit{loop-speedup}, which improves upon \textit{loop-original} by removing redundant or inefficient operations directly, acting like a strong substitute for the original design. 

\begin{figure}[bhtp]
	\begin{center}
	\includegraphics[width=1.0\columnwidth]
    {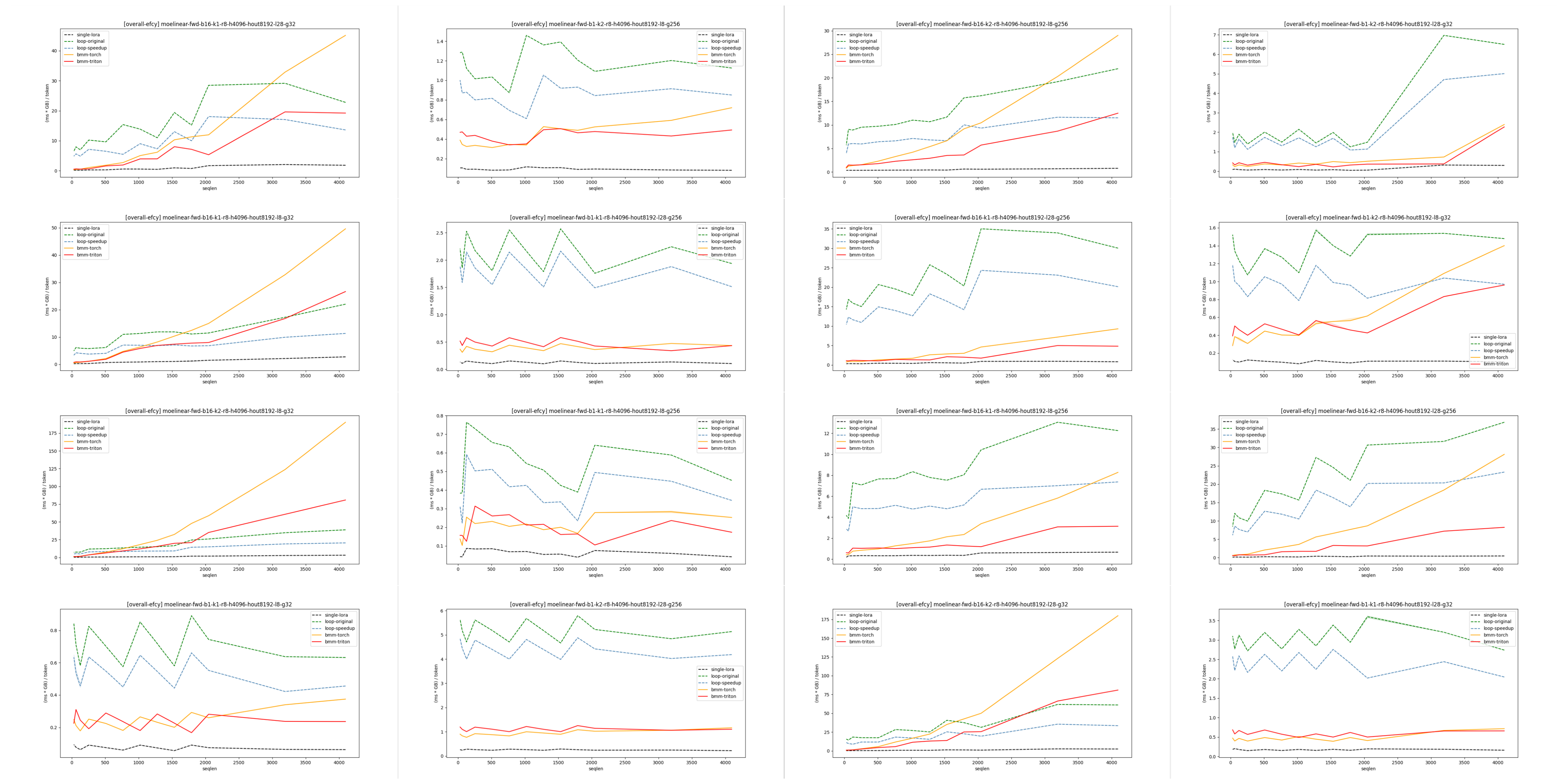}
		\caption{
        The overall efficiency evaluation curve displays the averaging runtime $\times$ memory footprint for each newly generated token (unit: ms $\times$ GB / token).
        \label{fig:overall-efcy_report_all}}
		\end{center}
\end{figure}

\begin{figure}[bhtp]
	\begin{center}
	\includegraphics[width=1.0\columnwidth]
    {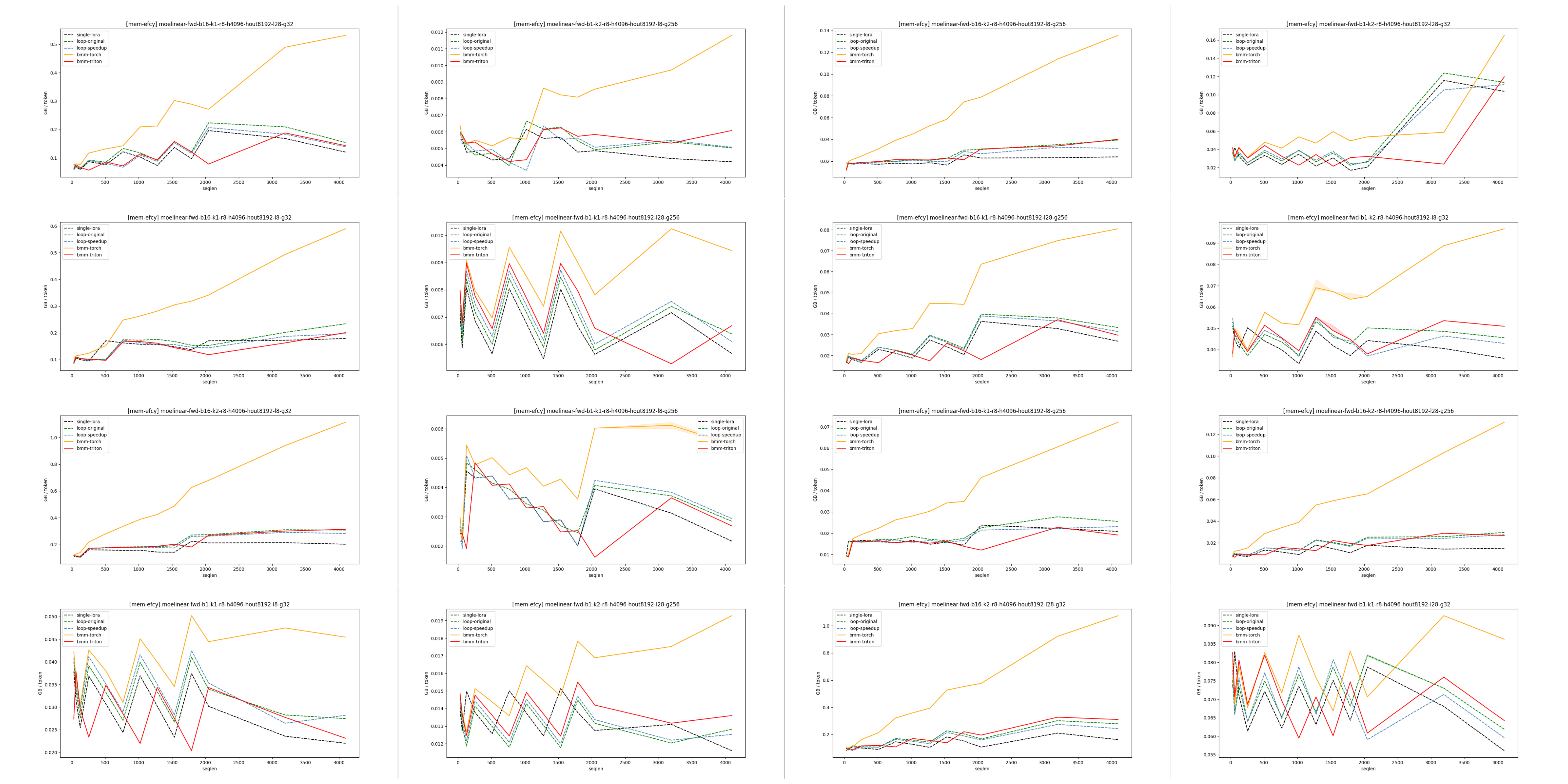}
		\caption{
        The memory efficiency evaluation curve displays the averaging memory footprint for each newly generated token (unit: GB / token).
        \label{fig:mem-efcy_report_all}}
		\end{center}
\end{figure}

\begin{figure}[thbp]
	\begin{center}
	\includegraphics[width=1.0\columnwidth]
    {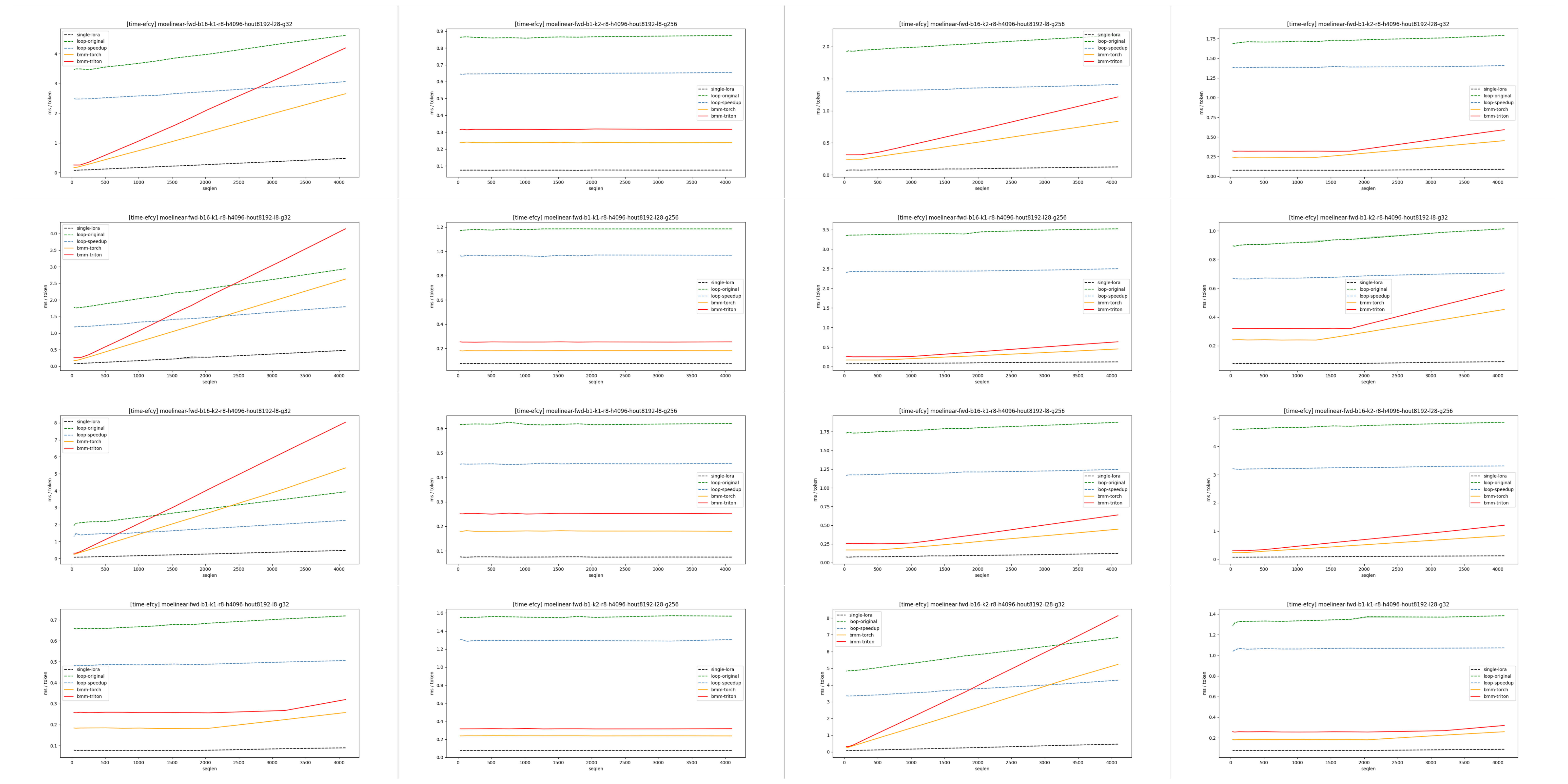}
		\caption{
        The time efficiency evaluation curve displays the averaging runtime for each newly generated token (unit: ms / token).
        \label{fig:time-efcy_report_all}}
		\end{center}
\end{figure}

As depicted in Figures~\ref{fig:mem-efcy_report_all} and~\ref{fig:time-efcy_report_all}, our \textit{bmm-torch} design outperforms other implementations, boasting an average speedup of $\sim\!4\times$ over \textit{loop-original}. However, its memory usage escalates with longer sequence lengths. In contrast, \textit{bmm-triton} maintains a comparable memory footprint to the baselines while retaining $80\!\%$ of the speedup achieved by \textit{bmm-torch}, showcasing a balanced trade-off between time and space, as illustrated in Figure~\ref{fig:overall-efcy_report_all}.

\newpage